\documentclass[runningheads]{llncs}
\usepackage{graphicx}
\usepackage{tikz}
\usepackage{comment}
\usepackage{amsmath,amssymb} 
\usepackage{color}

\usepackage{hhline}
\usepackage{colortbl}
\usepackage{multirow}
\usepackage[accsupp]{axessibility}  
\usepackage{orcidlink}
\usepackage{caption}

\usepackage{hyperref}

\newcommand\nnfootnote[1]{%
  \begin{NoHyper}
  \renewcommand\thefootnote{}\footnote{#1}%
  \addtocounter{footnote}{-1}%
  \end{NoHyper}
}

\definecolor{col1}{RGB}{243, 230, 195}
\definecolor{col2}{RGB}{208, 220, 242}
\definecolor{Gray}{gray}{0.9}
\newcommand{\etal}{\textit{et al.}}
\newcommand{\eg}{\textit{e.g.,}}

\begin{document}

\title{Disentangling Object Motion and Occlusion for Unsupervised Multi-frame Monocular Depth}

\titlerunning{Disentangling Object Motion and Occlusion for Unsupervised Multi-frame Monocular Depth}
%
\author{Ziyue Feng\inst{1}\orcidlink{0000-0002-0037-3697} \and
Liang Yang\inst{2}* \and
Longlong Jing\inst{2} \and
Haiyan Wang\inst{2} \and \\
YingLi Tian\inst{2} \and
Bing Li\inst{1}*}
\authorrunning{Z. Feng et al.}
\titlerunning{Dynamic Depth}
%
\institute{Clemson University, \and City University of New York}
\maketitle

\nnfootnote{*Corresponding authors \texttt{<bli4@clemson.edu>} \texttt{<lyang1@ccny.cuny.edu>}}
\begin{abstract}

Conventional self-supervised monocular depth prediction methods are based on a static environment assumption, which leads to accuracy degradation in dynamic scenes due to the mismatch and occlusion problems introduced by object motions. Existing dynamic-object-focused methods only partially solved the mismatch problem at the training loss level. In this paper, we accordingly propose a novel multi-frame monocular depth prediction method to solve these problems at both the prediction and supervision loss levels. Our method, called DynamicDepth, is a new framework trained via a self-supervised cycle consistent learning scheme. A Dynamic Object Motion Disentanglement (DOMD) module is proposed to disentangle object motions to solve the mismatch problem. Moreover, novel occlusion-aware Cost Volume and Re-projection Loss are designed to alleviate the occlusion effects of object motions. Extensive analyses and experiments on the Cityscapes and KITTI datasets show that our method significantly outperforms the state-of-the-art monocular depth prediction methods, especially in the areas of dynamic objects. Code is available at \url{https://github.com/AutoAILab/DynamicDepth}

\end{abstract}

\section{Introduction}

3D environmental information is crucial for autonomous vehicles, robots, and AR/VR applications. Self-supervised monocular depth prediction~\cite{monodepth,monodepth2,packnet,featdepth} provides an efficient solution to retrieve 3D information from a single camera without requiring expensive sensors or labeled data. In recent years these methods are getting more and more popular in both the research and industry communities.

Conventional self-supervised monocular depth prediction methods~\cite{monodepth,monodepth2,packnet} take a single image as input and predicts the dense depth map. They generally use a re-projection loss which constraints the geometric consistency between adjacent frames in the training loss level, but they are not capable of geometric reasoning through temporal frames in the network prediction level, which limits their overall performance.

\begin{figure*}
\centering
\includegraphics[width=.9\linewidth]{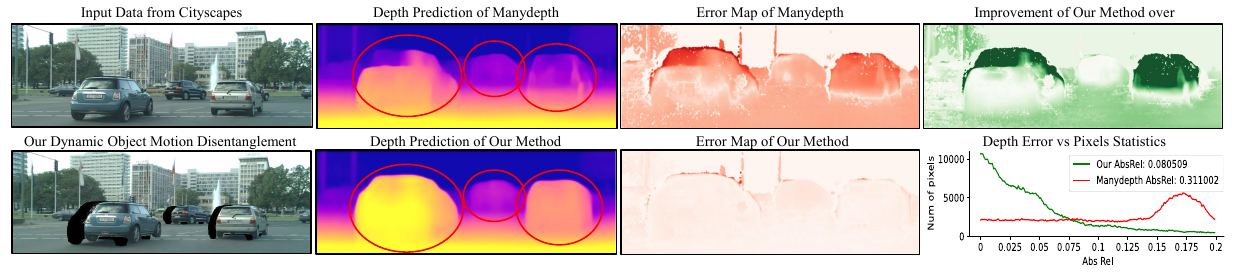}
\caption{  Conventional monocular depth prediction methods like Manydepth~\cite{manydepth} makes severe mistakes on dynamic object areas due to mismatch and occlusion problems introduced by object motions. Our method achieved significant improvement with our proposed Dynamic Object Motion Disentanglement and Occlusion Alleviation.}
\label{fig:1}
\end{figure*}

Temporal and spatially continuous images are available in most real-world scenarios like autonomous vehicles~\cite{Cityscapes,kitti} or smart devices~\cite{ha2016high,joshi2014micro}. Recent years multi-frame monocular depth prediction methods~\cite{cs2018depthnet,manydepth,monorec,patil2020dont,wang2019recurrent,zhang2019exploiting} are proposed to utilize the temporal image sequences to improve the depth prediction accuracy. Cost-volume-based methods~\cite{manydepth,monorec} adopted the cost volume from stereo match tasks to enable the geometric reasoning through temporal image sequences in the network prediction level, and achieved overall state-of-the-art depth prediction accuracy while not requiring time-consuming recurrent networks.

However, both the re-projection loss function and the cost volume construction are based on the static environment assumption, which does not hold for most real-world scenarios. Object motion will violate this assumption and cause re-projection mismatch and occlusion problems. The cost volume and loss values in the dynamic object areas are unable to reflect the quality of depth hypothesis and prediction, which will mislead the model training. 
Recent work~\cite{sgdepth,monodepth2,frozen,instadm} attempted to optimize depth prediction of dynamic object areas and achieved noticeable improvements, but they still have several drawbacks. (1) They only solve the mismatch problem at the loss function level, still cannot reason geometric constraints through temporal frames for dynamic objects, which limits its accuracy potential. (2) The occlusion problem introduced by object motions is still unsolved. (3) Redundant object motion prediction networks increased the model complexity and does not work for the motions of non-rigid objects.

Pursuing accurate and generic depth prediction, we propose DynamicDepth, a self-supervised temporal depth prediction framework that disentangles the dynamic object motions. First, we predict a depth prior from the target frame and project to the reference frames for an implicit estimation of object motion without rigidity assumption, which is later disentangled by our Dynamic Object Motion Disentanglement (DOMD) module. We then build a multi-frame occlusion-aware cost volume to encode the temporal geometric constraints for the final depth prediction. In the training level, we further propose a novel occlusion-aware re-projection loss to alleviate the occlusion from the object motions, and a novel cycle consistent learning scheme to enable the final depth prediction and the depth prior prediction to mutually improve each other. To summarize, our contributions are as follows:

\begin{itemize}
    \item We propose a novel Dynamic Object Motion Disentanglement (DOMD) module which leverages an initial depth prior prediction to solve the object motion mismatch problem in the final depth prediction.
    \item We devise a Dynamic Object Cycle Consistent training scheme to mutually reinforce the Prior Depth and the Final Depth prediction.
    \item We design an Occlusion-aware Cost Volume to enable geometric reasoning across temporal frames even in object motion occluded areas, and a novel Occlusion-aware Re-projection Loss to alleviate the motion occlusion problem in training supervision.
    \item Our method significantly outperforms existing state-of-the-art methods on the Cityscapes~\cite{Cityscapes} and KITTI~\cite{kitti} datasets.
\end{itemize}

\section{Related Work}
In this section, we review self-supervised depth prediction approaches relevant to our proposed method in the following three categories: (1) single-frame, (2) multi-frame, (3) dynamic-objects-optimized. 

\textbf{Self-supervised Single-frame Monocular Depth Prediction: }Due to the limited availability of labeled depth data, self-supervised monocular depth prediction methods~\cite{monodepth,monodepth2,featdepth,casser2018depth,li2020unsupervised,packnet} have become more and more popular. Monodepth2~\cite{monodepth2} set a benchmark for robust monocular depth, FeatDepth~\cite{featdepth} tried to improve the low-texture area depth prediction, and PackNet~\cite{packnet} explored a more effective network backbone. These self-supervised depth models generally take a single frame as input and predict the dense depth map. In the training stage, the temporally neighboring frames are projected to the current image plane by the predicted depth map. If the prediction is accurate, the re-projected images are supposed to be identical to the actual current frame image. The training is based on enforcing the re-projection photo-metric~\cite{ssim} consistency.

These methods provided a successful paradigm to learn the depth prediction without labeled data, but they have a major and common problem with dynamic objects: the re-projection loss function assumes the environment is static, which does not hold for real-world applications. When objects are moving, even if the prediction is perfect, the re-projected reference image will still not match the target frame image. The loss signal from the dynamic object areas will generate misleading gradients to degrade the model performance. In contrast, our proposed Dynamic Object Motion Disentanglement solves this mismatch problem and achieves superior accuracy, especially in the dynamic object areas.

\textbf{Multi-frame Monocular Depth Prediction: }The above mentioned re-projection loss only uses temporal constraints at the training loss function level. The model itself does not take any temporal information as input for reasoning, which limits its performance. One promising way to improve self-supervised monocular depth prediction is to leverage the temporal information in the input and prediction stage. Early works~\cite{cs2018depthnet,patil2020dont,wang2019recurrent,zhang2019exploiting} explored recurrent networks to process image sequences for monocular depth prediction. These recurrent models are computationally expensive and do not explicitly encode and reason geometric constraints in their prediction. Recently, Manydepth~\cite{manydepth} and MonoRec~\cite{monorec} adopt the cost volumes from stereo matching tasks to enable the geometric-based reasoning during inference. They project the reference frame feature map to the current image plane with multiple pre-defined depth hypothesises, whose difference to the current frame feature maps are stacked to form the cost volume. Hypothesises which are closer to the actual depth are supposed to have a lower value in the cost volume, while the entire cost volume is supposed to encode the inverse probability distribution of the actual depth value. With this integrated cost volume, they achieve great overall performance improvement while preserving real-time efficiency. 

However, the construction of the cost volume relies on the static environment assumption as well, which leads to catastrophic failure in the dynamic object area. They either circumvent this problem~\cite{monorec} or simply use a $L1$ loss~\cite{manydepth} to mimic the prediction of the single-frame model, which makes less severe mistakes for dynamic objects. This $L1$ loss alleviates but does not actually solve the problem. Our proposed Dynamic Object Motion Disentanglement, Occlusion-aware Cost Volume, and Re-projection Loss solve the mismatch and occlusion problem at both the reasoning and the training loss levels and outperform all other methods, especially in the dynamic object areas.

\textbf{Dynamic Objects in Self-supervised Depth Prediction: }The research community has attempted to solve the above-mentioned ill-posed re-projection geometry for dynamic objects. SGDepth~\cite{sgdepth} tried to exclude the moving objects from the loss function, Li \etal~\cite{frozen} proposed to build a dataset only containing non-moving dynamic-category objects. The latest state-of-the-art methods~\cite{casser2018depth,gao2020attentional,gordon2019depth,instadm,lee2021attentive,li2020unsupervised} tried to predict pixel-level or object-level translation and incorporate it into the loss function re-projection geometry. 

However, these methods still have several drawbacks. First, their single frame input did not enable the model to reason from the temporal domain. Second, explicitly predicting object motions requires redundant models and increased complexity. Third, they only focused on the re-projection mismatch, the occlusion problem introduced by object motions is still unsolved. Our proposed Dynamic Object Motion Disentanglement works at both the cost volume and the loss function levels, solving the re-projection mismatch problem while enabling the geometric reasoning through temporal frames in the inference stage, without additional explicit object motion prediction. Furthermore, we propose Occlusion-aware Cost Volume and Occlusion-aware Re-projection Loss to solve the occlusion problem introduced by object motion.

\section{Method}

\begin{figure*}[t]
\centering
\includegraphics[width=.9\linewidth]{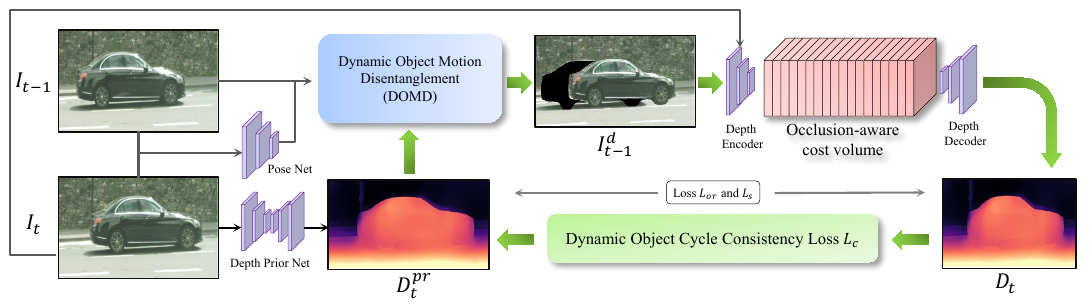}
   \caption{ \textbf{DynamicDepth Architecture:} The inputs are images $I_{t-1}$ and $I_{t}$, from which dynamic-object-disentangled frame $I_{t-1}^{d}$ is generated by the DOMD module for the final depth prediction $D_t$. The occlusion-aware cost volume is constructed to facilitate geometric reasoning and the Dynamic Object Cycle Consistency Loss is devised for mutual reinforcement between $D_t$ and $D^{pr}_t$. Green arrows indicates knowledge flow.}

\label{fig:overview}
\end{figure*}

\subsection{Overview}

Given two images $I_{t - 1} \in R^{W\times H \times 3}$ and $I_{t} \in R^{W\times H \times 3}$ of a target scene, our purpose is to estimate a dense depth map $D_t$ of $I_{t}$ by taking advantage of two views' observations while solving the mismatch and occlusion problems introduced by object motions.

As shown in Fig.~\ref{fig:overview}, our model contains three major innovations: We first use a Depth Prior Net $\theta_{DPN}$ and Pose Net $\theta_{p}$ to predict an initial depth prior $D^{pr}_t$ and ego-motion, which is sent to the (1) Dynamic Object Motion Disentanglement (DOMD) to solve the object motion mismatch problem (see Sec.~\ref{sec:domd}). The disentangled frame $I^{d}_{t-1}$ and the current frame $I_{t}$ are encoded by the Depth Encoder to construct the (2) Occlusion-aware Cost Volume for reasoning through temporal frames while diminishing the motion occlusion problem (see Sec.~\ref{sec:cv}). The final depth prediction $D_t$ is generated by the Depth Decoder from our cost volume. During training, our (3) Dynamic Object Cycle Consistency Loss $L_{c}$ enables the mutual improvement of the depth prior $D^{pr}_t$ and the final depth prediction $D_t$, while our Occlusion-aware Re-projection Loss $L_{or}$ solved the object motion occlusion problem (see Sec.~\ref{sec:loss}).

\subsection{Dynamic Object Motion Disentanglement (DOMD)}
\label{sec:domd}
There is an observation~\cite{casser2018depth,monodepth2} that single-frame monocular depth prediction models suffer from dynamic objects, which cause even more severe problems in multi-frame methods~\cite{manydepth,monorec}. This is because the static environment assumption does not hold for dynamic objects, which introduce mismatch and occlusion problems. Here, we describe our DOMD to solve the mismatch problem.

\subsubsection{Why the Cost Volume and Self-supervision Mismatch on Dynamic Objects:}
\begin{figure}[t]
\centering
\begin{minipage}{.7\textwidth}
  \centering
  \includegraphics[width=\linewidth]{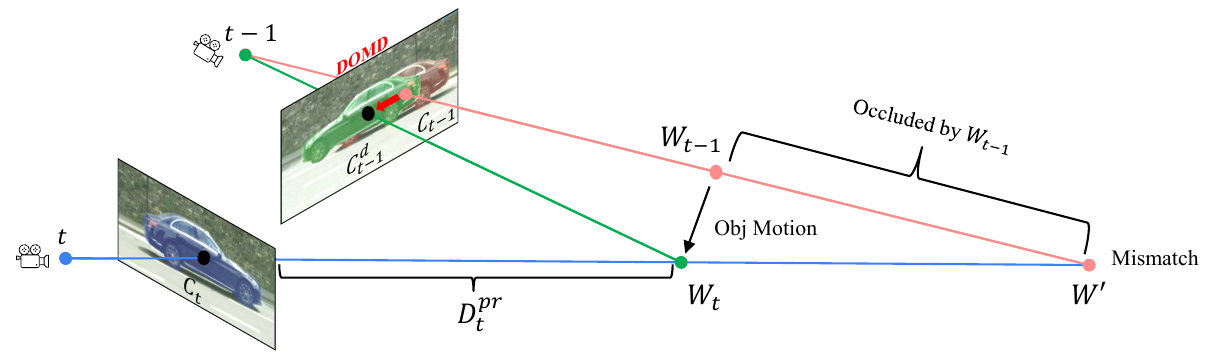}
  \setcaptionwidth{.95\textwidth}
  \caption{\textbf{Dynamic Object Motion Disentanglement:} A dynamic object moves from $W_{t-1}$ to $W_{t}$, $C_{t-1}$ and $C_{t}$ are corresponding image patches. $D^{pr}_t$ is our depth prior prediction. Conventional methods tend to mismatch at $W'$. We re-project $C_{t}$ to $C^{d}_{t-1}$ with depth prior $D^{pr}_t$ to replace $C_{t-1}$ to disentangle the object motion. This solves the mismatch problem, making our cost volume and re-projection loss correctly converge at $W_{t}$.}
  \label{fig:proj}
\end{minipage}%
\begin{minipage}{.3\textwidth}
  \centering
  \includegraphics[width=0.95\linewidth]{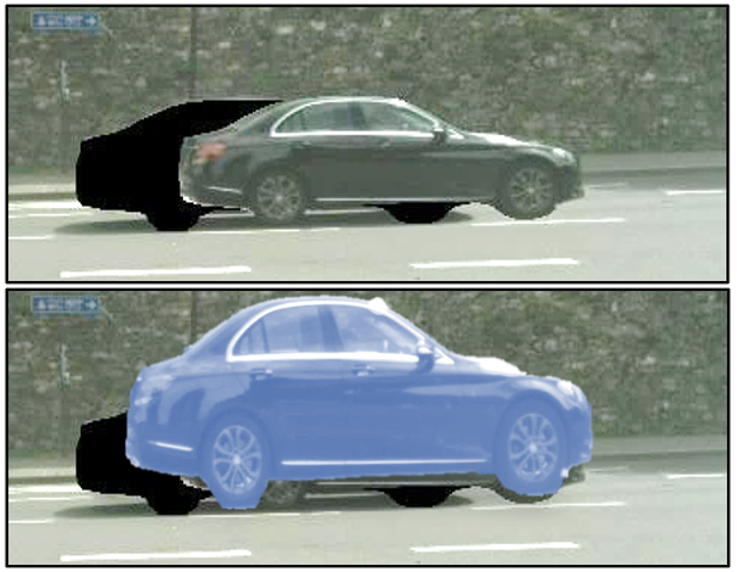}
  \caption{\textbf{Dynamic object motion disentangled image:} Left is the $I^{d}_{t-1}$ when depth prior is accurate. The right blue image patch shows the re-projected $C^{d}_{t-1}$ with inaccurate depth prior.}
  \label{fig:improve}
\end{minipage}
\end{figure}

Either in the cost volume or re-projection loss function, the current frame feature map $F_{t}$ or image $I_{t}$ is projected to the 3D space and re-projected back to the reference frame $t-1$ by the depth hypothesis or predictions. We illustrate the re-projection geometry in Fig.~\ref{fig:proj}. The dynamic object moves from $W_{t-1}$ to $W_{t}$, its corresponding image patches are $C_{t-1}$ and $C_{t}$ respectively. Conventional methods suppose the photo-metric difference between $C_{t-1}$ and the re-projected $C_{t}$ is lowest when the depth prediction or hypothesis is correctly close to $W_{t}$. However, due to the object motions, image or feature patches tend to mismatch at $W'$ instead: $E(C_{t-1}, \pi_{t-1}(W')) < E(C_{t-1}, \pi_{t-1}(W_{t}))$, $\pi$ is the projection operator. This mismatch misleads the reasoning in the cost volume and the supervision in the re-projection loss.


\subsubsection{Dynamic Object Motion Disentanglement: }
Our DOMD module $M_{o}$ takes two image frames ($I_{t-1}, I_{t}$) with its dynamic category (\eg vehicle, people, bike) segmentation masks ($S_{t-1}, S_{t}$) as input to generate the disentangled image $I^{d}_{t-1}$.
\begin{eqnarray}
\mathrm{M_{o}} : (I_{t}, I_{t-1}, S_{t-1}, S_{t}) \mapsto {I}^{d}_{t-1}.
\end{eqnarray}

We first use a single-frame depth prior network $\theta_{DPN}$ to predict an initial depth prior $D^{pr}_t$. As shown in Fig.~\ref{fig:proj}, the $D^{pr}_t$ is used to re-project the dynamic object image patch $C_{t}$ to $C^{d}_{t-1}$, which indicates the $t-1$ camera perspective of the dynamic object at location $W_{t}$. Finally, we replace the $C_{t-1}$ with $C^{d}_{t-1}$ to form the dynamic object motion disentangled image $I^{d}_{t-1}$. Note that we do not require the rigidity of the dynamic object.
\begin{eqnarray}
C_{a} = I_{a} \cdot S_{a},\quad
C^{d}_{t-1} = \pi_{t-1}(\pi_{t}^{-1}(C_{t}, D^{pr}_t)),\quad
I^{d}_{t-1} = I_{t-1}(C_{t-1} \to C^{d}_{t-1}).
\end{eqnarray}

Our Multi-frame model $\theta_{MF}$ then construct the geometric constraint in the cost volume with the disentangled image frame $I^{d}_{t-1}$ and current image frame $I_{t}$ to predict the final depth $D_t$.

We further propose a Dynamic Object Cycle Consistency Loss $L_{c}$ (Details in Sec.~\ref{sec:loss} and Sec.~\ref{sec:ablation}.) to enable the $D_t$ to backward supervise the $D^{pr}_t$ training. Both the $D^{pr}_t$ and $D_t$ could be greatly improved with our cycle consistent learning. Our $\theta_{DPN}$ already outperforms the existing dynamic-object-focused state-of-the-art methods such as InstaDM~\cite{instadm} with joint and cycle consistent learning.

\subsubsection{Why Final Depth Improves Over Depth Prior:}
As shown in Fig.~\ref{fig:improve}, when the depth prior prediction is inaccurate, the re-projected image patch $C^{d}_{t-1}$ will occlude some background pixels which are visible at time $t$. Those pixels will generate a higher photometric error in the re-projection loss. To minimize it, the network will manage to decode the error of depth prior from the disentangled image $I^{d}_{t-1}$ to predict a better final depth to improve the depth prior prediction by our later introduced cycle-consistency loss.

\subsection{Occlusion-aware Cost Volume}
\label{sec:cv}
\begin{figure*}[t]
\centering
\includegraphics[width=.9\linewidth]{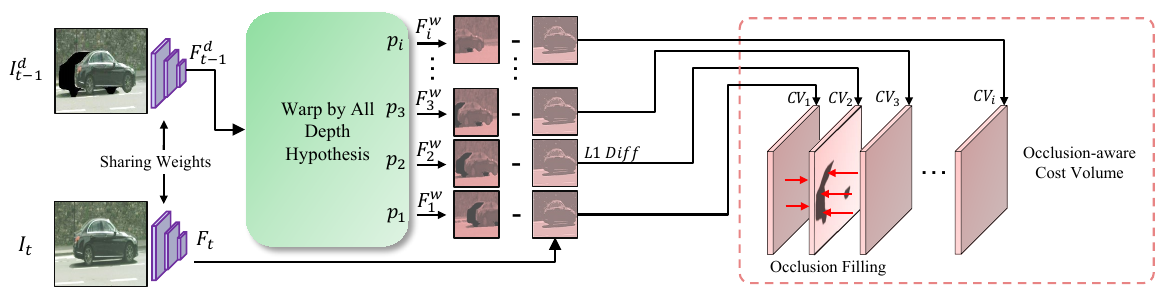}
\caption{ \textbf{Occlusion-aware Cost Volume: }Feature map $F^{d}_{t-1}$ of the $I^{d}_{t-1}$ is warped to the $I_{t}$ plane with multiple pre-defined depth hypothesizes $P_{i}$ to construct the cost volume. The black area in the cost volume indicates the noise from object motion occlusion, which is replaced with the nearby non-occluded area to avoid polluting the cost distribution.}
\label{fig:cv}
\end{figure*}
To encode the geometric constraints through the temporal frames while solving the occlusion problem introduced by dynamic objects motions, we propose an Occlusion-aware Cost Volume $CV^{occ} \in R^{|P|\times W\times H\times C}$, where $P=\{p_{1}, p_{2}, ..., p_{|P|}\}$ is the pre-defined depth hypothesis, $C$ is the channel number.

As shown in Fig.~\ref{fig:cv}, we warp the feature map $F^{d}_{t-1}$ of the dynamic object disentangled image $I^{d}_{t-1}$ to the current frame image plane with all pre-defined depth hypothesis $P$. The cost volume layer $CV_{i}$ is the $L1$ difference between the warped feature map $F^{w}_{i}$ and the current frame feature map $F_{t}$. We obtain the cost volume $CV$ by stacking all the layers. For each pixel, the cost value is supposed to be lower when the corresponding depth hypothesis is closer to the actual depth. The cost values over different depth hypotheses are supposed to encode the inverse probability distribution of the actual depth.
\begin{eqnarray}
CV_{i} = \left|F_{t} - F^{w}_{i} \right|_1, \quad \quad F^{w}_{i} = \pi_{t}(\pi_{t-1}^{-1}(F^{d}_{t-1}, p_{i})).
\end{eqnarray}

In Fig.~\ref{fig:cv}, the black area in the image $I^{d}_{t-1}$ corresponds to the backgrounds which may be visible at time $t$ but are occluded by the dynamic object at time $t-1$. The $L1$ difference between the feature of backgrounds at time $t$ and the feature of black pixels is meaningless, which pollutes the distribution of the cost volume. We propose to replace these values with non-occluded area cost values from neighboring depth hypothesis $p'$. This preserves the global cost distribution and leads the training gradients flow to the nearby non-occluded areas. Our ablation study in Sec.~\ref{sec:exp} confirms the effectiveness of our design.
\begin{eqnarray}
CV^{occ}_{p,w,h}=
\begin{cases}
CV_{p,w,h},  &\text{if } F^{w}_{p,w,h} \in V,\\
CV_{p',w,h}, &\text{if } F^{w}_{p,w,h} \in O, F^{w}_{p',w,h} \in V, p'\in r,
\end{cases}
\end{eqnarray}
where $O/V$ are the set of occluded/visible areas in $F^{w}$, $r$ is the neighbors of $p$.

\subsection{Loss Functions}
\label{sec:loss}
During the training of our framework, our proposed Occlusion-aware Re-projection Loss $L_{or}$ enforces the re-projection consistency between adjacent frames while alleviating the influence of the object-motion-caused occlusion problem. Our joint learning and novel Dynamic Object Cycle Consistency Loss $L_{c}$ further enables the depth prior prediction $D^{pr}_t$ and final depth prediction $D_t$ to mutually reinforce each other to achieve the best performance.

\subsubsection{Dynamic Object Cycle Consistency Loss:}
As shown in Fig.~\ref{fig:overview}, during the self-supervised learning, our initial depth prior prediction $D^{pr}_t$ is used in our Dynamic Object Motion Disentanglement (DOMD) module to produce the motion disentangled reference frame $I^{d}_{t-1}$ which is later encoded in our Occlusion-aware Cost Volume to guide the final depth prediction $D_t$. To enable the multi-frame final depth $D_t$ to backward guide the learning of single-frame depth prior $D^{pr}_t$ to achieve a mutual reinforcement scheme, we propose a novel Dynamic Object Cycle Consistency Loss $L_{c}$ to enforce the consistency between $D_t$ and $D^{pr}_t$.

Since only the dynamic objects area of $D^{pr}_t$ are employed in our DOMD module, we only apply the Dynamic Object Cycle Consistency Loss $L_{c}$ at these areas and only active when the inconsistency is large enough:
\begin{eqnarray}
A = \{i\in I_{t}| \frac{\left|D^{i}_t - D^{pr,i}_{t}\right|_1}{\mathrm{min}\{D^{i}_t, D^{pr,i}_{t}\}} > 1 \},\\
L_{c} = \frac{1}{\left|A \cap S \right|}\sum_{i\in (A\cap S)} \left|D^{i}_t - D^{pr,i}_{t}\right|_1.
\end{eqnarray}
Where $S$ is the semantic segmentation mask of dynamic category objects. 

\subsubsection{Occlusion-aware Re-projection Loss: }

\begin{figure*}[t]
\centering
\includegraphics[width=.9\linewidth]{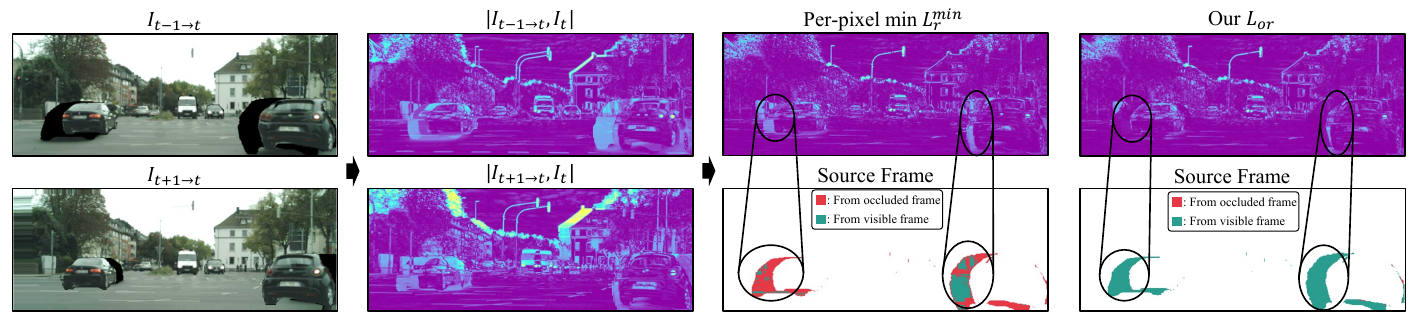}
\caption{ \textbf{Occlusion-aware Re-projection Loss: }Using the non-occluded source pixels for the re-projection loss could avoid most occlusions. The widely-used~\cite{monodepth2,manydepth,featdepth} per-pixel minimum $L^{min}_{r}$ fails when the occluded pixels do not have lower photo-metric error. We propose Occlusion-aware Re-projection Loss $L_{or}$ to solve this problem.}
\label{fig:lc}
\end{figure*}

In self-supervised monocular depth prediction, the image from reference frames ($I_{t-1}, I_{t+1}$) are warped to the current image plane with the predicted depth map $D_t$. If the depth prediction is correct, the conventional re-projection loss $L_{r}$ supposes the warped image $(I_{t-1 \to t}, I_{t+1 \to t})$ to be identical with the current frame image $I_{t}$. They penalize the photo-metric error $E$ between them.
\begin{eqnarray}
\hat E_{a} = E(I_{t}, I_{a \to t}), \qquad
L_{r} = \frac{1}{2} (\hat E_{t-1} + \hat E_{t+1}).
\end{eqnarray}

As mentioned above, the dynamic object motions break the static environment assumption and lead to the mismatch problem in this re-projection geometry. Our Dynamic Object Motion Disentanglement (DOMD) module $M_{o}$ could solve this mismatch problem but the background pixels occluded by the dynamic object at reference time ($t-1, t+1$) are still missing. As shown in Fig.~\ref{fig:lc}, using the photo-metric error $E$ between these occluded pixels in the warped image ($(I_{t-1 \to t}, I_{t+1 \to t})$) and visible background pixels in $I_{t}$ as training loss only introduces noise and misleads the model learning. 

Fortunately, object motions are normally consistent in a short time window, which means the backgrounds occluded at time $t-1$ are usually visible at time $t+1$ and vise-versa. It is possible to switch the source frame between $t-1$ and $t+1$ for each pixel to avoid the occlusion. The widely used per-pixel minimum re-projection loss~\cite{monodepth2} $L^{min}_{r}$ assumes the visible source pixels will have lower photo-metric error than the occluded ones, they thus proposed to choose the minimum error source frame for each pixel: $L^{min}_{r} = \frac{1}{|I_{t}|}\sum_{i\in I_{t}} min(\hat E^{i}_{t-1}, \hat E^{i}_{t+1})$.


However, in practice, as shown in the right columns of Fig.~\ref{fig:lc} we observe that around half of the visible source pixels do not have a lower photo-metric error than the occluded source. Since we can obtain the exact occlusion mask $O$ and visible mask $V$ from our DOMD module $M_{o}$, we propose Occlusion-aware Re-projection Loss $L_{or}$, which always choose the non-occluded source frame pixels for photo-metric error. More details are in the supplementary materials.

Following \cite{monodepth,peloss}, a combination of $L1$ norm and SSIM~\cite{ssim} with coefficient $\gamma$ is used as our photo-metric error $E_p$. The SSIM takes the pixels within a local window into account for error computation. In $I_{t-1\to t}$ and $I_{t+1\to t}$ the occluded pixels thus influence the neighboring non-occluded pixel's SSIM error. We propose Occlusion Masking $M_{a}$, which paints the corresponding pixels in target frame $I_{t}$ to be black when calculating the SSIM error with reference frames. This neutralizes the influence of the occlusion areas on neighboring pixels in SSIM. The ablation study in Sec.~\ref{sec:ablation} confirms applying our source pixel switching and occlusion masking mechanisms together makes the best improvement in the depth prediction quality.
\begin{eqnarray}
E_p\left[I_a, I_b\right] = \frac{\gamma}{2} (1 \!-\! \mathrm{SSIM}(I_a, I_b)) \!+\! (1 \!-\! \gamma) \left | I_a - I_b\right |_1 &.\\
EO_{t'} = E_p\left[M_{a}(I_{t}), I_{t' \to t}\right], \qquad \qquad \quad&
\end{eqnarray}

We further adopt the edge-aware metric from~\cite{ddvo} into our smoothness loss $L_{s}$ to make it invariant to output scale, which is formulated as:
\begin{eqnarray}
L_{s} &=& \left | \partial_x d^*_t \right | e^{-\left | \partial_x I_t \right |} + \left | \partial_y d^*_t   \right | e^{-\left | \partial_y I_t \right |},
\end{eqnarray} where $d^*_t = d_t / \overline{d_t}$ is the mean-normalized inverse depth, $\partial$ is the image gradient.

Our final loss $L$ is the sum of our Dynamic Object Cycle Consistency Loss $L_{c}$, Occlusion-aware Re-projection Loss $L_{or}$, and smoothness loss $L_{s}$:
\begin{eqnarray}
L = L_{c} + L_{or} + 1e^{-3}\cdot L_{s}.
\end{eqnarray}

\section{Experiments}
\label{sec:exp}

\begin{figure*}[t]
\centering
\includegraphics[width=.9\linewidth]{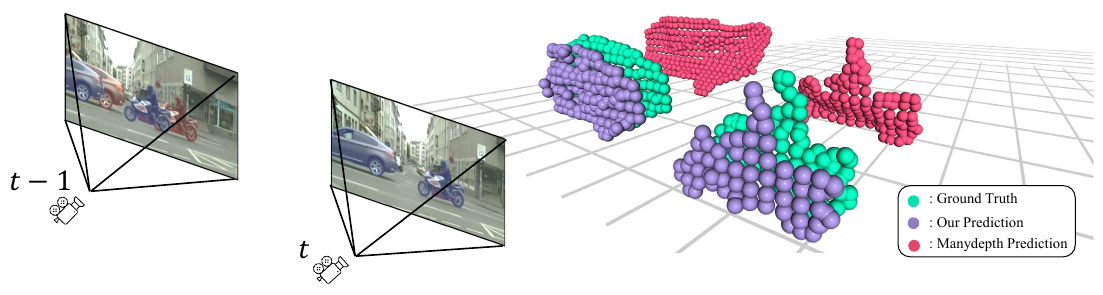}
\caption{\textbf{Error Visualization: }In the left $t-1$ image, {\color{red}red} image patch is the original data used by the Manydepth~\cite{manydepth} while the {\color{blue}blue} patch is generated by the DOMD module for our prediction. We project the dynamic object depths into point clouds. Our prediction matches the ground truth better.}
\label{fig:pcl}
\end{figure*}

\newcommand{\x}{ x }
\newcommand{\midline}{  }

\newcommand{\splitline}{\arrayrulecolor{black}\hhline{~------------}}

\newcommand*\rot{\rotatebox{90}}

\definecolor{Asectioncolor}{RGB}{255, 200, 200}
\definecolor{Bsectioncolor}{RGB}{255, 228, 196}
\definecolor{Csectioncolor}{RGB}{235, 255, 235}
\definecolor{Dsectioncolor}{RGB}{235, 235, 255}
\newcolumntype{a}{>{\columncolor{Gray}}c}

\begin{table*}[t]
  \centering
  \footnotesize
  \resizebox{.9\textwidth}{!}{
    \begin{tabular}{|l|l|c|c||a|c|c|c|c|c|c|}
        \arrayrulecolor{black}\hline
          &\multirow{2}*{Method} &\multirow{2}*{Test frames} &\multirow{2}*{WxH} &\multicolumn{4}{c|}{\cellcolor{col1}The lower the better} &\multicolumn{3}{c|}{\cellcolor{col2}The higher the better}\\
          ~& ~& ~ &~ &\cellcolor{col1}Abs Rel & \cellcolor{col1}Sq Rel & \cellcolor{col1}RMSE  & \cellcolor{col1}RMSE log & \cellcolor{col2}$\delta < 1.25 $ & \cellcolor{col2}$\delta < 1.25^{2}$ & \cellcolor{col2}$\delta < 1.25^{3}$ \\
         
        \hline\hline
        \parbox[b]{2mm}{\multirow{14}{*}{\rotatebox[origin=c]{90}{KITTI}}} 
        & Ranjan \etal \cite{ranjan2018adversarial}  & 1 & 832\x256 & 0.148 & 1.149 & 5.464 & 0.226 & 0.815 & 0.935 & 0.973\\
        \midline
        &  EPC++ \cite{luo2019every} & 1  & 832\x256 & 0.141 & 1.029 & 5.350 & 0.216 & 0.816 & 0.941 & 0.976\\
        \midline
        &  Struct2depth (M) \cite{casser2018depth}  & 1   &416\x128& 0.141 & 1.026 & 5.291 &  0.215 & 0.816 & 0.945 & {0.979}\\
        \midline
        & Li \etal \cite{li2020unsupervised}  & 1     & 416\x128 &0.130 & 0.950 & 5.138 & 0.209 & 0.843 & 0.948 & 0.978 \\
        \midline
        &  Videos in the wild \cite{gordon2019depth} & 1  &  416\x128& 0.128 & 0.959  & 5.230 & 0.212 & 0.845 & 0.947 & 0.976 \\
        \midline
        & Monodepth2 \cite{monodepth2} & 1  &  640\x192 & 0.115 &   0.903 &   4.863 &   0.193 &   0.877&   0.959 &   0.981 \\ 
        \midline
        & Lee \etal~\cite{lee2021attentive} & 1 & 832\x256 &0.114 &0.876 &4.715 &0.191 &0.872 &0.955 &0.981\\
        \midline
        & InstaDM~\cite{instadm}  & 1     & 832\x256 &0.112 &0.777 &4.772 &0.191 &0.872 &0.959 &0.982 \\
         \midline
        & Packnet-SFM \cite{packnet} & 1 & 640\x192 & 0.111 & 0.785 & 4.601 & 0.189 & 0.878 & 0.960 & 0.982 \\
        \midline
        & Johnston \etal~\cite{johnston2020self} & 1 & 640\x192 & 0.106 & 0.861 & 4.699 & 0.185 & 0.889 & 0.962 & 0.982 \\
        \midline
         &Guizilini \etal \cite{guizilini2020semantically} & 1  & 640\x192 &0.102 & 0.698 & 4.381 & 0.178 & 0.896 & 0.964 & 0.984 \\
        \midline
        & Patil \etal \cite{patil2020dont}  & N     & 640\x192  & 0.111  & 0.821  & 4.650  & 0.187  & 0.883  & 0.961  & 0.982 \\
        \midline
        & Wang \etal \cite{wang2020self} & 2 (-1, 0)     & 640\x192  & 0.106  & 0.799  & 4.662  & 0.187  & 0.889  & 0.961  & 0.982 \\
        \midline
        &ManyDepth~\cite{manydepth} & 2 (-1, 0)  & 640\x192 &   0.098  &   0.770  &   4.459  &   0.176  &   \textbf{0.900}  &   \textbf{0.965}  &   0.983 \\
        \midline
        &\textbf{DynamicDepth} & 2 (-1, 0)   & 640\x192 &    \textbf{0.096}  &   \textbf{0.720}  &  \textbf{4.458}  &  \textbf{0.175}  &  0.897  &  0.964  &  \textbf{0.984}\\
        
        \hline\hline

        \parbox[b]{2mm}{\multirow{10}{*}{\rotatebox[origin=c]{90}{Cityscapes}}} 
        \midline
        &Pilzer \etal \cite{pilzer2018unsupervised} & 1 &  512\x256 &0.240 & 4.264 & 8.049 & 0.334 & 0.710 &  0.871 & 0.937 \\
        &Struct2Depth 2 \cite{Casser_2019_CVPR_Workshops}  &  1  &   416\x128 &0.145  & 1.737  & 7.280  &  0.205 & 0.813 & 0.942 & 0.976 \\
        \midline
        &Monodepth2 \cite{monodepth2} & 1  & 416\x128 &0.129  &   1.569  &   6.876  &   0.187  &   0.849  &   0.957  &   0.983 \\
        \midline
        &Videos in the Wild \cite{gordon2019depth} & 1 &  416\x128 &{0.127} & {1.330} & {6.960} & {0.195} & {0.830} & {0.947} & {0.981} \\
        \midline
        &Li \etal \cite{li2020unsupervised} & 1 & 416\x128 &0.119 &  1.290 & {6.980} &  {0.190} & {0.846} &  0.952 &  0.982 \\
        \midline
        & Lee \etal~\cite{lee2021attentive} & 1 & 832\x256 &0.116 &1.213 &6.695 &0.186 &0.852 &0.951 &0.982\\
        \midline
        & InstaDM~\cite{instadm}  & 1     & 832\x256 &0.111 &1.158 &6.437 &0.182 &0.868 &0.961 &0.983 \\
        \midline
        &Struct2Depth 2 \cite{Casser_2019_CVPR_Workshops}  &  3 (-1,  0,  +1)  & 416\x128 &0.151 & 2.492 & 7.024 & 0.202 & 0.826 & 0.937 & 0.972 \\
        \midline
        &ManyDepth~\cite{manydepth} & 2 (-1, 0)  &  416\x128 &0.114  &   1.193  &   6.223  &   0.170  &   0.875  &   0.967  &   0.989 \\
        \midline
        &\textbf{DynamicDepth} & 2 (-1, 0)   & 416\x128 &\textbf{0.103} &\textbf{1.000} &\textbf{5.867} &\textbf{0.157} &\textbf{0.895} &\textbf{0.974} &\textbf{0.991}\\
         
        \arrayrulecolor{black}\hline

    \end{tabular}
  } 
  \caption{\footnotesize \textbf{Depth Prediction on KITTI and Cityscapes Dataset.} Following the convention, methods in each category are sorted by the Abs Rel, which is the relative error with the ground truth. Best methods are in \textbf{bold}. Our method out-performs all other state-of-the-art methods by a large margin especially on the challenging Cityscapes~\cite{Cityscapes} dataset, which contains significantly more dynamic objects. Note that all KITTI result in this table are based on the widely-used original~\cite{kitti} dataset, which generates much greater error than the improved~\cite{kitti2017} dataset.
    }

\label{tab:overall}
\end{table*}

The experiments are mainly focused on the challenging Cityscapes~\cite{Cityscapes} dataset, which contains many dynamic objects. To comprehensively compare with more state-of-the-art methods, we also report the performance on the widely-used KITTI~\cite{kitti} dataset. Since our method is mainly focused on the dynamic objects, we further conduct additional evaluation on the depth errors of the dynamic objects areas, which clearly demonstrate the effectiveness of our method. The design decision and the effectiveness of our proposed framework is evaluated by an extensive ablation study.

\subsection{Implementation Details: }We use frames $\{I_{t-1}, I_{t}, I_{t+1}\}$ for training and $\{I_{t-1}, I_{t}\}$ for testing. All dynamic objects is this paper are determined by an off-the-shelf semantic segmentation model EffcientPS~\cite{efficientps}. Note that we do not need instance-level masks and inter-frame correspondences, all dynamic category pixels are projected together at once. All network modules including the depth prior net $\theta_{DPN}$ are trained together from scratch or ImageNet~\cite{imagenet} pre-training. ResNet18~\cite{resnet} is used as the backbone. We use the Adam~\cite{adam} optimizer with a learning rate of $10^{-4}$ to train for 10 epochs, which takes about $10$ hours on a single Nvidia A100 GPU.


 
  

\begin{table*}[t]
  \centering

  \resizebox{.9\textwidth}{!}{
  \begin{tabular}{|c|l|c||a|c|c|c|c|c|c|}
\arrayrulecolor{black}\hline
&\multirow{2}*{Method} &\multirow{2}*{WxH} &\multicolumn{4}{c|}{\cellcolor{col1}The lower the better} &\multicolumn{3}{c|}{\cellcolor{col2}The higher the better}\\
~& ~ &~ &\cellcolor{col1}Abs Rel & \cellcolor{col1}Sq Rel & \cellcolor{col1}RMSE  & \cellcolor{col1}RMSE log & \cellcolor{col2}$\delta < 1.25 $ & \cellcolor{col2}$\delta < 1.25^{2}$ & \cellcolor{col2}$\delta < 1.25^{3}$ \\
\hline\hline
\parbox[b]{2mm}{\multirow{5}{*}{\rotatebox[origin=c]{90}{KITTI}}} 
    
&Monodepth2~\cite{monodepth2}   & 640\x192 &0.169  &1.878  &5.711  &0.271  &0.805  &0.909  &0.944\\

&InstaDM~\cite{instadm}         & 832\x256  &\underline{0.151}  &\underline{1.314}  &5.546  &0.271  &0.805  &0.905  &0.946\\

&ManyDepth~\cite{manydepth}     & 640\x192  &0.175  &2.000  &5.830  &0.278  &0.776  &0.895  &0.943\\

&\textbf{Our Depth Prior}       & 640\x192  &0.155  &1.317  &\underline{5.253}  &\underline{0.269}  &\underline{0.805}  &\underline{0.908}  &\underline{0.946}\\

&\textbf{DynamicDepth}          & 640\x192  &\textbf{0.150}  &\textbf{1.313}  &\textbf{5.146}  &\textbf{0.264}  &\textbf{0.807}  &\textbf{0.915}  &\textbf{0.949}\\
\hline\hline
\parbox[b]{2mm}{\multirow{5}{*}{\rotatebox[origin=c]{90}{Cityscapes}}} 
&Monodepth2~\cite{monodepth2}   & 416\x128  &0.159  &1.937  &6.363  &0.201  &0.816  &0.950  &0.981\\

&InstaDM~\cite{instadm}         & 832\x256  &0.139  &1.698  &5.760  &0.181  &\underline{0.859}  &0.959  &0.982\\

&ManyDepth~\cite{manydepth}     & 416\x128  &0.169  &2.175  &6.634  &0.218  &0.789  &0.921  &0.969\\

&\textbf{Our Depth Prior}       & 416\x128  &\underline{0.137}  &\underline{1.285}  &\underline{4.674}  &\underline{0.174}  &0.852  &\underline{0.961}  &\underline{0.985}\\

&\textbf{DynamicDepth}   &  416\x128 &\textbf{0.129}  &\textbf{1.273}  &\textbf{4.626}  &\textbf{0.168}  &\textbf{0.862}  &\textbf{0.965}  &\textbf{0.986} \\

    \arrayrulecolor{black}\hline
  \end{tabular}
  }
  \caption{\textbf{Depth Error on Dynamic Objects.} We evaluate the depth prediction errors of dynamic objects (\eg Vehicles, Person, Bike) on KITTI~\cite{kitti} and Cityscapes~\cite{Cityscapes} datasets. The best results are in \textbf{bold}, second best are \underline{underlined}. Our depth prior prediction $D^{pr}_t$ already outperform the state-of-the-art method InstaDM~\cite{instadm} using the same single frame input, while our final depth prediction $D_t$ sets a new benchmark.}
  
  \label{tab:doj}
\end{table*}

\subsection{Cityscapes Results}
 Cityscapes~\cite{Cityscapes}  is a challenging dataset with significant amount of dynamic objects. It contains $5,000$ videos each with $30$ frames, totaling  $150,000$ image frames. We exclude the first, last, and static-camera frames in each video for training, resulting in $58,335$ frames training data. The official testing set contains $1,525$ image frames. 

\begin{figure*}[t]
\centering
\includegraphics[width=.9\linewidth]{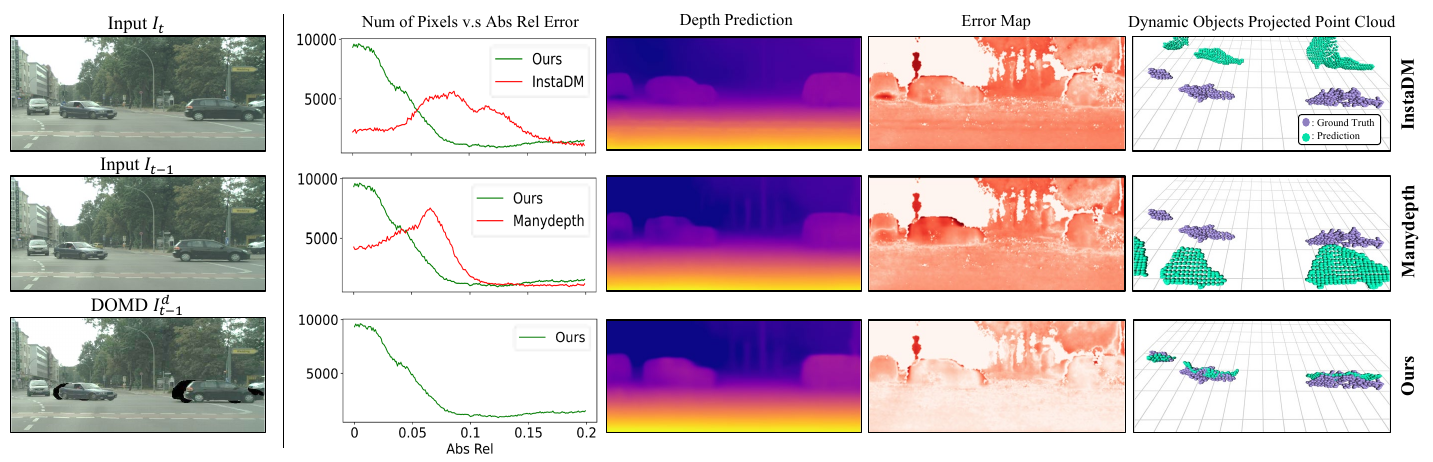}
\caption{ \textbf{Qualitative visualization: }The left column shows the input image frames and our disentangled image $I^{d}_{t-1}$, later columns show the comparison with other state-of-the-art methods. In the Histograms, most pixels of our method has lower depth error. In the error map, our method has lighter {\color{red}red} color which indicates lower depth errors. We project the dynamic object area depths to 3D point clouds and compare them with ground truth point clouds in the last column. Our prediction matches the ground truth significantly better. More comparisons are provided in the supplementary document.}
\label{fig:ql}
\end{figure*}

Table~\ref{tab:overall} shows the depth prediction results on the Cityscapes~\cite{Cityscapes} testing set. Following the convention, we rank all methods based on the absolute-relative-errors. Since the Cityscapes dataset contains significant amount of dynamic objects, the object-motion-optimized method InstaDM~\cite{instadm} achieved the best accuracy among all the existing methods. With the help of our proposed Dynamic Object Motion Disentanglement (DOMD), Dynamic Object Cycle Consistency Loss, Occlusion-aware Cost Volume and the Occlusion-aware Re-projection Loss, our method outperforms the InstaDM~\cite{instadm} by a large margin in all of the metrics using a lower resolution and more concise architecture (we do not require the explicit per-object-motion network, instance level segmentation prior and inter-frame correspondences). Qualitative visualizations are in Fig.~\ref{fig:ql}.

Table~\ref{tab:doj} shows the depth errors in the dynamic objects area. Our Depth Prior Network $\theta_{DPN}$ shares a similar architecture with the Monodepth2~\cite{monodepth2} while trained jointly with our multi-frame model $\theta_{MF}$ using Dynamic Object Cycle Consistency Loss $L_{c}$. It outperforms all the existing methods including Monodepth2~\cite{monodepth2} and InstaDM~\cite{instadm}. Manydepth~\cite{manydepth} suffers catastrophic failure on the dynamic objects due to the aforementioned mismatch and occlusion problems. They employed an separate single-frame model as a teacher for dynamic objects area. However, since it does not actually solve the mismatch and occlusion problems, it still makes severe mistakes on dynamic objects. In contrast, with our proposed innovations, our multi-frame model $\theta_{MF}$ boosts up the accuracy even higher, achieves superior advantages on all the metrics, showing its significant effectiveness. We show a qualitative visualization in Fig.~\ref{fig:pcl}.

\subsection{KITTI Results}
Our proposed framework is further evaluated on the widely-used KITTI~\cite{kitti} dataset Eigen~\cite{eigen-split} split, which contains $39,810$ training images, $4,424$ validation images, and $697$ testing images. According to our statistic, only 0.34\% of the pixels in the KITTI~\cite{kitti} dataset are dynamic category objects (\eg Vehicle, Person, Bike), and most of the vehicles are not moving. 

The comparison of our method with the state-of-the-art single-frame models~\cite{monodepth2,casser2018depth,li2020unsupervised,packnet}, multi-frame models~\cite{patil2020dont,wang2020self,manydepth}, and dynamic-objects-optimized models~\cite{instadm,lee2021attentive} is summarized in Table~\ref{tab:overall}. Unsurprisingly dynamic-objects-focused methods~\cite{casser2018depth,instadm,lee2021attentive,gao2020attentional,gordon2019depth,li2020unsupervised} showed a minor advantage on this dataset. Our method only achieve 2\% improvement over the existing state-of-the-art method Manydepth~\cite{manydepth}. However, when we only focus on dynamic objects as in Table~\ref{tab:doj}, our method achieve a much more significant 14.3\% improvement.

\begin{table*}[t]
\centering
\resizebox{.9\textwidth}{!}{
\begin{tabular}{|c|c|c|c|c|cccc|}
\hline
\cellcolor{col2}Dynamic Object &\cellcolor{col2}Dynamic Object &\cellcolor{col2}Occlusion-aware &\multicolumn{2}{c|}{\cellcolor{col2}Occlusion-aware Loss} &\multicolumn{4}{c|}{\cellcolor{col1}The Lower the Better}\\
\cellcolor{col2}Motion Disentanglement&\cellcolor{col2}Cycle Consistency &\cellcolor{col2}Cost Volume&\cellcolor{col2}Switching &\cellcolor{col2}Masking &\cellcolor{col1}Abs Rel &\cellcolor{col1}Sq Rel &\cellcolor{col1}RMSE &\cellcolor{col1}RMSE$_{log}$ \\
\hline
\multicolumn{9}{|c|}{\cellcolor{Gray}Evaluating Dynamic Object Motion Disentanglement}\\
\hline
           &           &           &           &           &0.114   &   1.193  &   6.223  &   0.170 \\
\hline
\checkmark &           &           &           &           &0.110   &   1.172  &   6.220  &   0.166\\
\hline 
\multicolumn{9}{|c|}{\cellcolor{Gray}Evaluating Occlusion-aware $CV$ and Loss}\\
\hline
\checkmark &           &           &           &           &0.110   &   1.172  &   6.220  &   0.166\\
\hline 
\checkmark &           &           &\checkmark &           &0.110   &   1.168  &   6.223  &   0.166\\
\hline
\checkmark &           &           &           &\checkmark &0.110   &   1.167  &   6.210  &   0.167\\
\hline
\checkmark &           &           &\checkmark &\checkmark &0.108   &   1.139  &   5.992  &   0.163\\
\hline
\checkmark &           &\checkmark &           &           &0.108   &   1.131  &   5.994  &   0.162\\
\hline
\multicolumn{9}{|c|}{\cellcolor{Gray}Evaluating Dynamic Object Cycle Consistent Training}\\
\hline
\checkmark &           &\checkmark &\checkmark &\checkmark &0.107   &   1.121  &   5.924  &   0.160\\
\hline
\checkmark &\checkmark &\checkmark &\checkmark &\checkmark &\textbf{0.103} &\textbf{1.000} &\textbf{5.867} &\textbf{0.157}\\
\hline
\end{tabular}}
\caption{\textbf{Ablation Study: }
Evaluating the effects for our proposed Dynamic Object Motion Disentanglement, Cycle Consistent Training, Occlusion-aware Cost Volume and Re-projection Loss on the Cityscapes~\cite{Cityscapes} dataset.
}
\label{tab:ablation}
\end{table*}

\subsection{Ablation Study}
\label{sec:ablation}

To comprehensively understand the effectiveness of our proposed modules and prove our design decision, we perform an extensive ablation study on the challenging Cityscapes~\cite{Cityscapes} dataset. As shown in Table~\ref{tab:ablation}, our experiments fall into three groups, evaluating Dynamic Object Motion Disentanglement, Occlusion-aware Cost Volume and Loss, and Cycle Consistent Training.

\textbf{Dynamic Object Motion Disentanglement:}
In the first group of the Table~\ref{tab:ablation}, we evaluate our proposed Dynamic Object Motion Disentanglement (DOMD) module. When the DOMD is enabled, the cost volume and the re-projection loss is based on the disentangled $I^{d}_{t-1}$ image instead of the original $I_{t-1}$ image. The Abs Rel Error reduced by 4\%, confirms its effectiveness.

\textbf{Occlusion-aware Cost Volume and Loss:}
The second group of the Table~\ref{tab:ablation} shows the effectiveness of the proposed Occlusion-aware Cost Volume $CV^{occ}$ and Occlusion-aware Re-projection Loss $L_{or}$. Our innovation in the Occlusion-aware Re-projection Loss includes two operations: the switching and masking. Solely using either the switching or masking mechanism does not improve the accuracy. These results meet our expectation. The re-projection loss switching mechanism is designed to switch the re-projection source between two reference frames $I^{d}_{t-1}$ and $I^{d}_{t+1}$ to avoid occlusion areas, and the masking mechanism is designed to neutralize the influence on the photo-metric error~\cite{ssim} from occlusion areas to neighboring non-occluded areas. Only avoiding the occlusion area while ignoring its influence on the neighboring areas or vise-versa could not solve the problem. Applying both mechanisms together can significantly improve the depth accuracy. As for the Occlusion-aware Cost Volume, our occlusion-filling mechanism replaces the noisy occluded cost voxels with neighboring non-occluded voxel values to recover the distribution of the costs and guide the training gradients. Experiments confirm the effectiveness of our design.

\textbf{Cycle Consistent training:}
The depth prior prediction $D^{pr}_t$ from $\theta_{DPN}$ is used in our DOMD module to disentangle the dynamic objects motion, which is further encoded with geometric constraints in the cost volume to predict the final depth $D_t$. The proposed Dynamic Object Cycle Consistency Loss $L_{c}$ enables the final depth $D_t$ to backwards supervise the training of the depth prior prediction $D^{pr}_t$ and forms a closed-loop mutual reinforcement. 
In the first row of the Table~\ref{tab:ablation} third group, we first train the Depth Prior Net $\theta_{DPN}$ separately, then freeze it and train the later multi-frame model to cut off the backwards supervision. In this experiment, $\theta_{DPN}$ performs similar as normal single-frame model Monodepth2~\cite{monodepth2} and the final depth prediction only shows limited performance. In the last row, when we unfreeze the $\theta_{DPN}$ to enable the joint and consistent training, our model achieves the best performance.

\section{Conclusions}
We presented a novel self-supervised multi-frame monocular depth prediction model, namely DynamicDepth. It disentangle object motions and diminish occlusion effects caused by dynamic objects, achieved the state-of-the-art performance especially at the dynamic object areas on the Cityscapes~\cite{Cityscapes} and KITTI~\cite{kitti} datasets. 

\textbf{Acknowledgement: }
This work was partially supported by the U.S. Department of Transportation (DOT) Center for Connected Multimodal Mobility grant \# No. 69A3551747117-2024230, and National Science Foundation (NSF) grant \# No. IIS-2041307.


\clearpage

\begin{center}
    \huge
    \textbf{\underline{Supplementary Materials}}
    \\[20pt]
    \normalsize 
    \small
\end{center}
\setcounter{section}{0}

\section{Additional Implementation Details}
\textbf{Occlusion-aware Re-projection Loss:} 
We obtain the exact occlusion mask $O$ and visible mask $V$ from our DOMD module $M_{o}$, our Occlusion-aware Re-projection Loss $L_{or}$ always choose the non-occluded source frame pixels for photo-metric error.
\begin{eqnarray}
L_{or} = \frac{1}{|I_{t} - (O_{t-1} \cap O_{t+1})|}\sum_{i\in I_{t}} E^{i}_{or},\qquad \quad&\\
E^{i}_{or} = 
\begin{cases}
EO^i_{t-1}, & \text{if } I_{t-1}^{i} \in V_{t-1}, I_{t+1}^{i} \in O_{t+1}, \\
EO^i_{t+1}, & \text{if } I_{t-1}^{i} \in O_{t-1}, I_{t+1}^{i} \in V_{t+1}, \\
\mathrm{min}(EO^{i}_{t-1}, EO^{i}_{t+1}), & \text{if } I_{t-1}^{i} \in V_{t-1}, I_{t+1}^{i} \in V_{t+1}, \\
0, & \text{if } I_{t-1}^{i} \in O_{t-1}, I_{t+1}^{i} \in O_{t+1}. \\
\end{cases}
\end{eqnarray}

\textbf{Depth Prior Net:} Our Depth Prior Net $\theta_{DPN}$ consists of a depth encoder and a depth decoder. We use an ImageNet~\cite{imagenet} pre-trained ResNet18~\cite{resnet} as backbone for depth encoder, which has 4 pyramidal scales. Features in each scale are fed to the depth decoder by several UNet~\cite{unet} style skip connections. The depth decoder consists of multiple convolution layers for the encoder feature fusion and nearest interpolations for up-sampling.

\textbf{Pose Net:} Our Pose Net shares a similar architecture as our Depth Prior Net, but it outputs a 6-degree-of-freedom camera ego-motion vector $P_{o}$ instead of the depth map.

\textbf{DOMD:} Our Dynamic Object Motion Disentanglement (DOMD) module projects the object image patches $C_{t}$ to $C^{d}_{t-1}$ to replace $C_{t-1}$ to disentangle the object motion. The projection is based on the depth prior prediction $D^{pr}_t$, known camera intrinsics $K$, and camera ego-motion prediction $P_{o}$. We do not need instance-level masks and inter-frame correspondences, all dynamic objects are projected together at once. We use an off-the-shelf semantic segmentation model EffcientPS~\cite{efficientps} to provide the dynamic category segmentation masks. We define the dynamic category as follows: \{person, rider, car, truck, bus, caravan, trailer, motorcycle, bicycle\}.

\textbf{Cost Volume:} We pre-define $96$ different depth hypothesis bins and reduce the channel number to $1$. The cost volume is constructed at the third scale which is in $48\times 160$ resolution, resulting in an $CV \in R^{96\times160\times48\times1}$.  Our cost volume only consumes $2.8MB$ memory when using Float32 data type.

\textbf{Depth Encoder and Decoder:} Our depth encoder and decoder in the multi-frame model $\theta_{MF}$ shares the same architecture with the Depth Prior Net $\theta_{DPN}$. The Occlusion-aware Cost Volume is integrated at the third scale of the encoder.

\textbf{Training:} We use frames $\{I_{t-1}, I_{t}, I_{t+1}\}$ for training and $\{I_{t-1}, I_{t}\}$ for testing. Our model is trained using an Adam~\cite{adam} optimizer with a learning rate of $10^{-4}$ for 10 epochs, which takes about $10$ hours on a single Nvidia A100 GPU.

\textbf{Evaluation Metrics:} Following the state-of-the-art methods~\cite{monodepth2,packnet,featdepth}, we use Absolute Relative Error (Abs Rel), Squared Relative Error (Sq Rel), Root Mean Squared Error (RMSE), Root Mean Squared Log Error (RMSE$_{log}$), and $\delta_1$, $\delta_2$, $\delta_3$ as the metrics to evaluate the depth prediction performance. These metrics are formulated as:

$
  \mathrm{Abs Rel}= \frac{1}{n} \sum_{i} \frac{|p_i - g_i|}{g_i},\qquad \qquad \qquad
  \mathrm{Sq Rel}=  \frac{1}{n} \sum_{i} \frac{(p_i - g_i)^2}{g_i},
 $
 
 $
  \mathrm{RMSE}= \sqrt{\frac{1}{n} \sum_{i}(p_i - g_i)^2},\quad \quad
  \mathrm{RMSE_{log}}=  \sqrt{\frac{1}{n} \sum_{i}(\log p_i - \log  g_i)^2},
  $
  
  $
  \delta_1, \delta_2, \delta_3= \%\ of\ thresh < 1.25, 1.25^2, 1.25^3,
$
where $g$ and $p$ are the depth values of ground truth and prediction in meters, $thresh=\max(\frac{g}{p},\frac{p}{g})$.

\section{Additional Quantitative Results}

\subsection{KITTI Benchmark Scores}
The original Eigen~\cite{eigen-split} split of KITTI~\cite{kitti} dataset uses the re-projected single-frame raw LIDAR points as ground truth for evaluation, which may contain outliers such as reflection on transparent objects. We only reported results with this original ground truth in the main paper since it is the most widely used. Jonas \etal~\cite{kitti2017} introduced a set of high-quality ground truth depth maps for the KITTI dataset, accumulates 5 consecutive frames to form the denser ground truth depth map, and removed the outliers. This improved ground truth depth is provided for $652$ (or $93\%$) of the $697$ test frames contained in the Eigen test split~\cite{eigen-split}. We evaluate our method on these $652$ improved ground truth frames and compare with existing state-of-art published methods in Table~\ref{tab:ktimp}. Following the convention, we clip the predicted depths to 80 meters to match the Eigen evaluation. Methods are ranked by the Absolute Relative Error. Our method outperforms all existing state-of-the-art methods, even some stereo-based and supervised methods.

\subsection{Full Quantitative Results}
Due to the space limitation, we only show a part of the quantitative comparison of depth prediction in the main paper. Here we show an extensive comparison to existing state-of-the-art methods on the KITTI~\cite{kitti} and Cityscapes~\cite{Cityscapes} dataset in Table.~\ref{tab:overallfull}. Following the convention, methods are sorted by the Abs Rel, which is the relative error with the ground truth. Our method outperforms all other state-of-the-art methods by a large margin, especially on the challenging Cityscapes~\cite{Cityscapes} dataset, which contains significantly more dynamic objects. Our method even outperformed some stereo-based and supervised methods on the KITTI dataset. Note that all KITTI results in this section are based on the widely-used original~\cite{kitti} ground truth, which generates much greater error than the improved~\cite{kitti2017} ground truth.
\section{Additional Qualitative Results}
\begin{figure*}[t]
\centering
\includegraphics[width=1\linewidth]{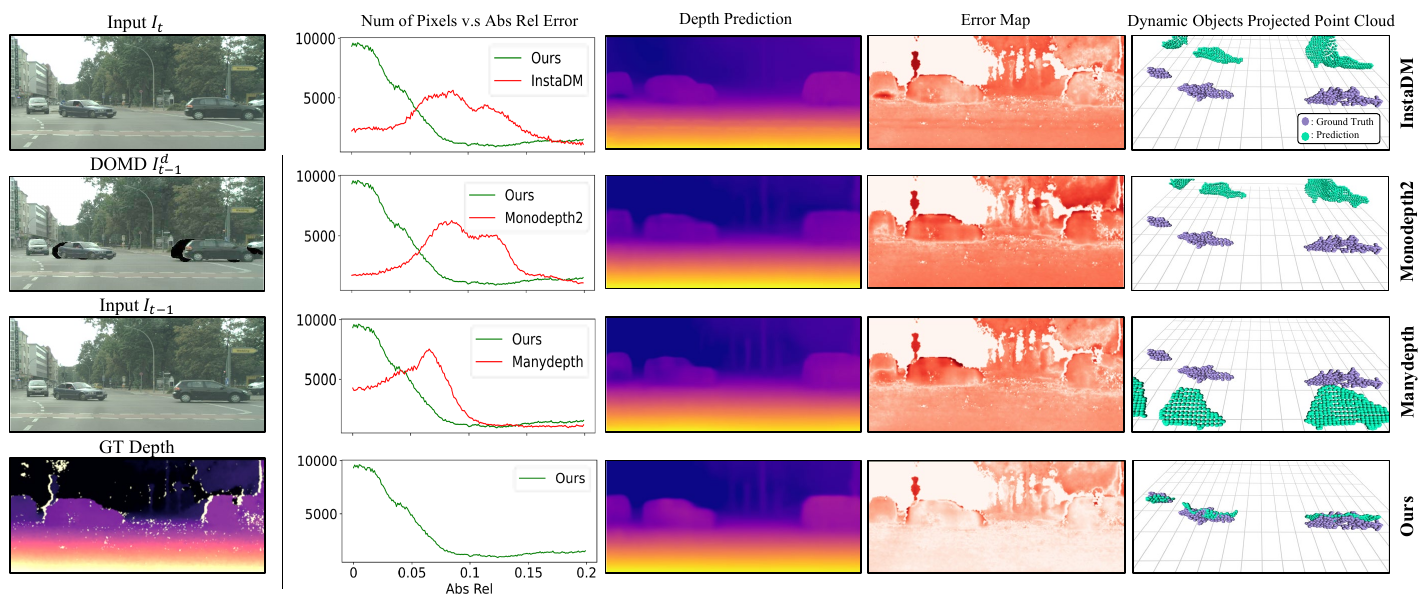}
\caption{ \textbf{Full Qualitative visualization: }The left column shows the input image frames and our disentangled image $I^{d}_{t-1}$, later columns show the comparison with other state-of-the-art methods. In the histograms, most pixels of our method has lower depth error. In the error map, our method has lighter {\color{red}red} color which indicates lower depth errors. We project the dynamic object area depths to 3D point clouds and compare them with ground truth point clouds in the last column. Our prediction matches the ground truth significantly better.}
\label{fig:ql1}
\end{figure*}

\begin{figure*}[t]
\centering
\includegraphics[width=1\linewidth]{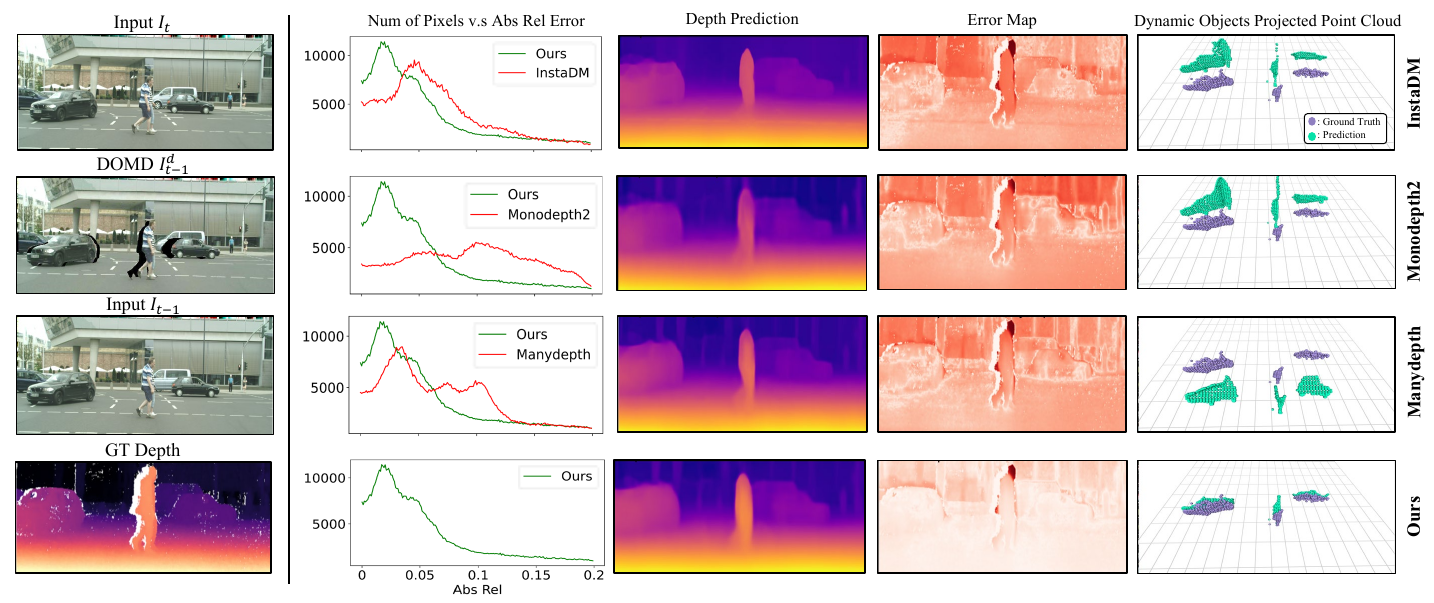}
\caption{ \textbf{Additional Qualitative visualization: }The left column shows the input image frames and our disentangled image $I^{d}_{t-1}$, later columns show the comparison with other state-of-the-art methods. In the histograms, most pixels of our method has lower depth error. In the error map, our method has lighter {\color{red}red} color which indicates lower depth errors. We project the dynamic object area depths to 3D point clouds and compare them with ground truth point clouds in the last column. Our prediction matches the ground truth significantly better.}
\label{fig:ql2}
\end{figure*}

Fig~\ref{fig:ql1} shows a full version of the qualitative results and Fig~\ref{fig:ql2} shows an additional set of comparisons. We compare our results with other state-of-the-art methods. The $I^{d}_{t-1}$ image disentangled the dynamic object motion to solve the mismatch problem. As shown in the histograms, most pixels of our method have lower depth error. Our method has lighter {\color{red}red} color in the error map which indicates lower depth errors. The dynamic object area depths are projected to 3D point clouds and compared with ground truth point clouds, our prediction matches the ground truth significantly better.

\begin{table*}[t]
  \centering
  
  \resizebox{1.0\textwidth}{!}{
    \begin{tabular}{|l|c|c||a|c|c|c|c|c|c|}
\arrayrulecolor{black}\hline
\multirow{2}*{Method} &\multirow{2}*{Training} &\multirow{2}*{WxH} &\multicolumn{4}{c|}{\cellcolor{col1}The lower the better} &\multicolumn{3}{c|}{\cellcolor{col2}The higher the better}\\
~& ~ &~ &\cellcolor{col1}Abs Rel & \cellcolor{col1}Sq Rel & \cellcolor{col1}RMSE  & \cellcolor{col1}RMSE log & \cellcolor{col2}$\delta < 1.25 $ & \cellcolor{col2}$\delta < 1.25^{2}$ & \cellcolor{col2}$\delta < 1.25^{3}$ \\

\hline
Zhan FullNYU \cite{zhanst2018} & Sup  &608\x 160 & 0.130 & 1.520 & 5.184 & 0.205 & 0.859 & 0.955 & 0.981\\ 
Kuznietsov et al.~\cite{kuznietsov2017semi} &Sup &621\x 187 &0.089 &0.478 &3.610 &0.138 &0.906 &0.980 &0.995\\
DORN~\cite{dorn} &Sup &513\x 385 &0.072 &0.307 &2.727 &0.120 &0.932 &0.984 &0.995\\
\hline
Monodepth \cite{monodepth} & S &512\x 256 & 0.109 & 0.811 & 4.568 & 0.166 & 0.877 & 0.967 & 0.988\\
3net \cite{poggi20183net} (VGG) & S  &512\x 256 &   0.119 &   0.920 &   4.824 &   0.182 &   0.856 &   0.957 &   0.985 \\ 
3net \cite{poggi20183net} (ResNet 50) & S  &512\x 256 &   0.102 &   0.675 &   4.293 &   0.159 &   0.881 &   0.969 &0.991 \\ 
SuperDepth \cite{pillai2018superdepth} & S &1024\x 384 &0.090 &0.542 &3.967 &0.144 &0.901 &0.976 &0.993 \\
Monodepth2~\cite{monodepth2} & S &640\x192 &0.085  &0.537  &3.868  &0.139  &0.912  &0.979  &0.993  \\
EPC++ \cite{luo2019every} & S & 832\x256 & 0.123 & 0.754 & 4.453 & 0.172 & 0.863 & 0.964 &0.989\\

\hline
SfMLearner \cite{zhou2017unsupervised} & M &416\x128 & 0.176 & 1.532 & 6.129 & 0.244 & 0.758 & 0.921 & 0.971\\
Vid2Depth \cite{mahjourian2018unsupervised} & M &416\x128 & 0.134 & 0.983 & 5.501 & 0.203 & 0.827 & 0.944 & 0.981\\
GeoNet \cite{geonet2018} & M  &416\x128 & 0.132 & 0.994 & 5.240 & 0.193 & 0.833 & 0.953 & 0.985\\
DDVO \cite{wang2017learning} & M  &416\x128 & 0.126 & 0.866 & 4.932 & 0.185 & 0.851 & 0.958 & 0.986\\ 
Ranjan \cite{ranjan2018adversarial}  & M & 832\x256  &0.123 & 0.881 & 4.834 & 0.181 & 0.860 & 0.959 & 0.985\\
EPC++ \cite{luo2019every} & M  & 832\x256 &0.120 & 0.789 &  4.755  & 0.177  & 0.856 &  0.961 &  0.987\\
Johnston \etal~\cite{johnston2020self} & M & 640\x192 &0.081 & 0.484  & 3.716  & 0.126  & 0.927  & 0.985  & 0.996 \\
Monodepth2 \cite{monodepth2} & M  &  640\x192 &0.090 & 0.545 & 3.942 & 0.137 & 0.914 & 0.983 & 0.995\\ 
Packnet-SFM \cite{packnet} & M & 640\x192 &0.078 & 0.420 &  3.485 &  0.121 &  0.931 &  0.986 &  0.996\\
Patil \etal \cite{patil2020dont}  & M  & 640\x192&   0.087  &   0.495  &   3.775  &   0.133  &   0.917  &   0.983  &   0.995  \\
Wang \etal \cite{wang2020self} & M    & 640\x192  &0.082 & 0.462 & 3.739 & 0.127 &  0.923 &  0.984 & 0.996\\
ManyDepth~\cite{manydepth} & M  & 640\x192 &0.070  &0.399  &3.455  &0.113  &0.941  &0.989  &0.997\\

\textbf{DynamicDepth} & M   & 640\x192 &    \textbf{0.068}  &   \textbf{0.362}  &  \textbf{3.454}  &  \textbf{0.111}  &  \textbf{0.943}  &\textbf{0.991}  &  \textbf{0.998}\\
\arrayrulecolor{black}\hline

\end{tabular}
  } 
  \vspace{1mm}
  \caption{ \textbf{KITTI Evaluation on Improved Ground Truth~\cite{kitti2017}:} Following the convention, methods in each category are sorted by the Abs Rel, which is the relative error with the ground truth. Best methods are in \textbf{bold}. Our method out-performs all other state-of-the-art methods, even some stereo-based and supervised methods.
  {  
        \scriptsize
        \newline
        \textbf{Legend:} \hspace{10pt} 
        Sup -- Supervised by ground truth depth\hspace{10pt}
        S -- Stereo\hspace{10pt}
        M -- Monocular
    } 
    }

\label{tab:ktimp}
\end{table*}

\begin{table*}[t]
  \centering
  
  \resizebox{1.0\textwidth}{!}{
    \begin{tabular}{|l|l|c|c||a|c|c|c|c|c|c|}
        \arrayrulecolor{black}\hline
&\multirow{2}*{Method} &\multirow{2}*{Training} &\multirow{2}*{WxH} &\multicolumn{4}{c|}{\cellcolor{col1}The lower the better} &\multicolumn{3}{c|}{\cellcolor{col2}The higher the better}\\
~& ~& ~ &~ &\cellcolor{col1}Abs Rel & \cellcolor{col1}Sq Rel & \cellcolor{col1}RMSE  & \cellcolor{col1}RMSE log & \cellcolor{col2}$\delta < 1.25 $ & \cellcolor{col2}$\delta < 1.25^{2}$ & \cellcolor{col2}$\delta < 1.25^{3}$ \\
         
\hline\hline
\parbox[b]{2mm}{\multirow{38}{*}{\rotatebox[origin=c]{90}{KITTI Original}}} 
&Zhan FullNYU \cite{zhanst2018}       &Sup  &608\x 160 &0.135 &1.132 &5.585 &0.229 &0.820 &0.933 &0.971\\
&Kuznietsov et al.~\cite{kuznietsov2017semi} &Sup &621\x 187 &0.113 &0.741 &4.621 &0.189 &0.862 &0.960 &0.986\\
&Gur et al.~\cite{gur2019single}     &Sup  &416\x128 &0.110 &0.666 &4.186 &0.168 &0.880 &0.966 &0.988\\
&Dorn~\cite{dorn}                   &Sup &513\x 385 &0.099 &0.593 &3.714 &0.161 &0.897 &0.966 &0.986\\

&MonoDepth~\cite{monodepth}         &S &512\x256 &0.133 &1.142 &5.533 &0.230 &0.830 &0.936 &0.970\\
&MonoDispNet~\cite{monodispnet}     &S &512\x256 &0.126 &0.832 &4.172 &0.217 &0.840 &0.941 &0.973\\
&MonoResMatch~\cite{monoresmatch}   &S &1280\x384 &0.111 &0.867 &4.714 &0.199 &0.864 &0.954 &0.979\\
&MonoDepth2~\cite{monodepth2}       &S &640\x192 &0.107 &0.849 &4.764 &0.201 &0.874 &0.953 &0.977\\
&UnDeepVO~\cite{undeepvo}           &S &512\x128 &0.183 &1.730 &6.570 &0.268 &- &- &-\\
&DFR~\cite{dfr}                     &S &608\x160 &0.135 &1.132 &5.585 &0.229 &0.820 &0.933 &0.971\\
&EPC++~\cite{luo2019every}          &S & 832\x256 &0.128 &0.935 &5.011 &0.209 &0.831 &0.945 &0.979\\
&DepthHint~\cite{depthhint}         &S &640\x192 &0.100 &0.728 &4.469 &0.185 &0.885 &0.962 &0.982\\ 
&FeatDepth~\cite{featdepth}        &S &640\x192 &0.099 &0.697 &4.427 &0.184 &0.889 &0.963 &0.982\\
& SfMLearner~\cite{zhou2017unsupervised}&M &416\x128 &0.208 &1.768 &6.958 &0.283 &0.678 &0.885 &0.957\\
& Vid2Depth~\cite{mahjourian2018unsupervised}          &M &416\x128 &0.163 &1.240 &6.220 &0.250 &0.762 &0.916 &0.968\\
& LEGO~\cite{lego}                   &M &416\x128 &0.162 &1.352 &6.276 &0.252 &0.783 &0.921 &0.969\\
& GeoNet~\cite{geonet2018}           &M &416\x128 &0.155 &1.296 &5.857 &0.233 &0.793 &0.931 &0.973\\
& DDVO~\cite{ddvo}                   &M &416\x128 &0.151 &1.257 &5.583 &0.228 &0.810 &0.936 &0.974\\
& DF-Net~\cite{dfnet}                &M &576\x 160 &0.150 &1.124 &5.507 &0.223 &0.806 &0.933 &0.973\\
& Ranjan \etal \cite{ranjan2018adversarial}   & M & 832\x256 & 0.148 & 1.149 & 5.464 & 0.226 & 0.815 & 0.935 & 0.973\\
&  EPC++ \cite{luo2019every}                  & M  & 832\x256 & 0.141 & 1.029 & 5.350 & 0.216 & 0.816 & 0.941 & 0.976\\
&  Struct2depth (M) \cite{casser2018depth}    & M   &416\x128& 0.141 & 1.026 & 5.291 &  0.215 & 0.816 & 0.945 & {0.979}\\
&SIGNet~\cite{signet}                         & M & 416\x128 &0.133 &0.905 &5.181 &0.208 &0.825 &0.947 &0.981\\
& Li \etal \cite{li2020unsupervised}          & M    & 416\x128 &0.130 & 0.950 & 5.138 & 0.209 & 0.843 & 0.948 & 0.978 \\
&  Videos in the wild \cite{gordon2019depth}  & M  &  416\x128& 0.128 & 0.959  & 5.230 & 0.212 & 0.845 & 0.947 & 0.976 \\
&DualNet~\cite{dualnet}                       &M &1248\x 384 &0.121 &0.837 &4.945 &0.197 &0.853 &0.955 &0.982\\
&SuperDepth~\cite{pillai2018superdepth}       &M &1024\x 384 &0.116 &1.055 &-     &0.209 &0.853 &0.948 &0.977\\
& Monodepth2 \cite{monodepth2}                & M  &640\x192 & 0.115 &   0.903 &   4.863 &0.193 & 0.877 &0.959 &0.981 \\ 
& Lee \etal~\cite{lee2021attentive}           & M & 832\x256 &0.114 &0.876 &4.715 &0.191 &0.872 &0.955 &0.981\\
& InstaDM~\cite{instadm}                      & M    & 832\x256 &0.112 &0.777 &4.772 &0.191 &0.872 &0.959 &0.982 \\
& Patil \etal \cite{patil2020dont}            & M & 640\x192  & 0.111  & 0.821  & 4.650  & 0.187  & 0.883  & 0.961  & 0.982 \\
& Packnet-SFM \cite{packnet}                  & M & 640\x192 & 0.111 & 0.785 & 4.601 & 0.189 & 0.878 & 0.960 & 0.982 \\
& Wang \etal \cite{wang2020self}              & M & 640\x192  & 0.106  & 0.799  & 4.662  & 0.187  & 0.889  & 0.961  & 0.982 \\
& Johnston \etal~\cite{johnston2020self}      & M & 640\x192 & 0.106 & 0.861 & 4.699 & 0.185 & 0.889 & 0.962 & 0.982 \\
&FeatDepth~\cite{featdepth}                   &M &640\x192 &0.104 &0.729 &4.481 &0.179 &0.893 &0.965 &0.984\\
&Guizilini \etal \cite{guizilini2020semantically} & M  & 640\x192 &0.102 & 0.698 & 4.381 & 0.178 & 0.896 & 0.964 & 0.984 \\
&ManyDepth~\cite{manydepth}                   & M  & 640\x192 &   0.098  &   0.770  &   4.459  &   0.176  &   \textbf{0.900}  &   \textbf{0.965}  &   0.983 \\
&\textbf{DynamicDepth} & M   & 640\x192 &    \textbf{0.096}  &   \textbf{0.720}  &  \textbf{4.458}  &  \textbf{0.175}  &  0.897  &  0.964  &  \textbf{0.984}\\
        
\hline\hline

\parbox[b]{2mm}{\multirow{10}{*}{\rotatebox[origin=c]{90}{Cityscapes}}} 
&Pilzer \etal \cite{pilzer2018unsupervised} & M &  512\x256 &0.240 & 4.264 & 8.049 & 0.334 & 0.710 &  0.871 & 0.937 \\
&Struct2Depth 2 \cite{Casser_2019_CVPR_Workshops}  &  M  &   416\x128 &0.145  & 1.737  & 7.280  &  0.205 & 0.813 & 0.942 & 0.976 \\
&Monodepth2 \cite{monodepth2} & M  & 416\x128 &0.129  &   1.569  &   6.876  &   0.187  &   0.849  &   0.957  &   0.983 \\
&Videos in the Wild \cite{gordon2019depth} & M &  416\x128 &{0.127} & {1.330} & {6.960} & {0.195} & {0.830} & {0.947} & {0.981} \\
&Li \etal \cite{li2020unsupervised} & M & 416\x128 &0.119 &  1.290 & {6.980} &  {0.190} & {0.846} &  0.952 &  0.982 \\
& Lee \etal~\cite{lee2021attentive} & M & 832\x256 &0.116 &1.213 &6.695 &0.186 &0.852 &0.951 &0.982\\
& InstaDM~\cite{instadm}  & M     & 832\x256 &0.111 &1.158 &6.437 &0.182 &0.868 &0.961 &0.983 \\
&Struct2Depth 2 \cite{Casser_2019_CVPR_Workshops}  &  M  & 416\x128 &0.151 & 2.492 & 7.024 & 0.202 & 0.826 & 0.937 & 0.972 \\
&ManyDepth~\cite{manydepth} & M  &  416\x128 &0.114  &   1.193  &   6.223  &   0.170  &   0.875  &   0.967  &   0.989 \\
&\textbf{DynamicDepth} & M   & 416\x128 &\textbf{0.103} &\textbf{1.000} &\textbf{5.867} &\textbf{0.157} &\textbf{0.895} &\textbf{0.974} &\textbf{0.991}\\
         
\arrayrulecolor{black}\hline
\end{tabular}
 } 
  \vspace{1mm}
  \caption{ \textbf{Depth Prediction on KITTI and Cityscapes Dataset.} Following the convention, methods in each category are sorted by the Abs Rel, which is the relative error with the ground truth. Best methods are in \textbf{bold}. Our method out-performs all other state-of-the-art methods by a large margin especially on the challenging Cityscapes~\cite{Cityscapes} dataset, which contains significantly more dynamic objects. Our method even outperformed some stereo based and supervised methods on KITTI dataset. Note that all KITTI result in this table are based on the widely-used original~\cite{kitti} ground truth, which generates much greater error than the improved~\cite{kitti2017} ground truth.
  {  
        \scriptsize
        \newline
        \textbf{Legend:} \hspace{10pt} 
        Sup -- Supervised by ground truth depth\hspace{10pt}
        S -- Stereo\hspace{10pt}
        M -- Monocular
    } 
    }
    \vspace{-5mm}

\label{tab:overallfull}
\end{table*}

\clearpage
%
%
\bibliographystyle{splncs04}
\bibliography{egbib}

\begin{thebibliography}{10}
\providecommand{\url}[1]{\texttt{#1}}
\providecommand{\urlprefix}{URL }
\providecommand{\doi}[1]{https://doi.org/#1}

\bibitem{casser2018depth}
Casser, V., Pirk, S., Mahjourian, R., Angelova, A.: Depth prediction without
  the sensors: Leveraging structure for unsupervised learning from monocular
  videos. In: AAAI (2019)

\bibitem{Casser_2019_CVPR_Workshops}
Casser, V., Pirk, S., Mahjourian, R., Angelova, A.: Unsupervised monocular
  depth and ego-motion learning with structure and semantics. In: CVPR
  Workshops (2019)

\bibitem{Cityscapes}
Cordts, M., Omran, M., Ramos, S., Rehfeld, T., Enzweiler, M., Benenson, R.,
  Franke, U., Roth, S., Schiele, B.: The cityscapes dataset for semantic urban
  scene understanding. In: Proc. of the IEEE Conference on Computer Vision and
  Pattern Recognition (CVPR) (2016)

\bibitem{cs2018depthnet}
CS~Kumar, A., Bhandarkar, S.M., Prasad, M.: Depthnet: A recurrent neural
  network architecture for monocular depth prediction. In: Proceedings of the
  IEEE Conference on Computer Vision and Pattern Recognition Workshops. pp.
  283--291 (2018)

\bibitem{imagenet}
Deng, J., Dong, W., Socher, R., Li, L.J., Li, K., Fei-Fei, L.: Imagenet: A
  large-scale hierarchical image database. In: 2009 IEEE conference on computer
  vision and pattern recognition. pp. 248--255. Ieee (2009)

\bibitem{eigen-split}
Eigen, D., Fergus, R.: Predicting depth, surface normals and semantic labels
  with a common multi-scale convolutional architecture. In: Proceedings of the
  IEEE international conference on computer vision. pp. 2650--2658 (2015)

\bibitem{dorn}
Fu, H., Gong, M., Wang, C., Batmanghelich, K., Tao, D.: Deep ordinal regression
  network for monocular depth estimation. In: Proceedings of the IEEE
  conference on computer vision and pattern recognition. pp. 2002--2011 (2018)

\bibitem{gao2020attentional}
Gao, F., Yu, J., Shen, H., Wang, Y., Yang, H.: Attentional
  separation-and-aggregation network for self-supervised depth-pose learning in
  dynamic scenes. arXiv preprint arXiv:2011.09369  (2020)

\bibitem{monodepth}
Godard, C., Mac~Aodha, O., Brostow, G.J.: Unsupervised monocular depth
  estimation with left-right consistency. In: Proceedings of the IEEE
  conference on computer vision and pattern recognition. pp. 270--279 (2017)

\bibitem{monodepth2}
Godard, C., Mac~Aodha, O., Firman, M., Brostow, G.J.: Digging into
  self-supervised monocular depth estimation. In: Proceedings of the IEEE/CVF
  International Conference on Computer Vision. pp. 3828--3838 (2019)

\bibitem{gordon2019depth}
Gordon, A., Li, H., Jonschkowski, R., Angelova, A.: Depth from videos in the
  wild: Unsupervised monocular depth learning from unknown cameras. In: ICCV
  (2019)

\bibitem{packnet}
Guizilini, V., Ambrus, R., Pillai, S., Raventos, A., Gaidon, A.: {3D} packing
  for self-supervised monocular depth estimation. In: CVPR (2020)

\bibitem{guizilini2020semantically}
Guizilini, V., Hou, R., Li, J., Ambrus, R., Gaidon, A.: Semantically-guided
  representation learning for self-supervised monocular depth. In: ICLR (2020)

\bibitem{gur2019single}
Gur, S., Wolf, L.: Single image depth estimation trained via depth from defocus
  cues. In: Proceedings of the IEEE/CVF Conference on Computer Vision and
  Pattern Recognition. pp. 7683--7692 (2019)

\bibitem{ha2016high}
Ha, H., Im, S., Park, J., Jeon, H.G., Kweon, I.S.: High-quality depth from
  uncalibrated small motion clip. In: Proceedings of the IEEE conference on
  computer vision and pattern Recognition. pp. 5413--5421 (2016)

\bibitem{resnet}
He, K., Zhang, X., Ren, S., Sun, J.: Deep residual learning for image
  recognition. In: Proceedings of the IEEE conference on computer vision and
  pattern recognition. pp. 770--778 (2016)

\bibitem{johnston2020self}
Johnston, A., Carneiro, G.: Self-supervised monocular trained depth estimation
  using self-attention and discrete disparity volume. In: CVPR (2020)

\bibitem{joshi2014micro}
Joshi, N., Zitnick, C.L.: Micro-baseline stereo. Microsoft Research Technical
  Report  (2014)

\bibitem{adam}
Kingma, D.P., Ba, J.: Adam: A method for stochastic optimization. arXiv
  preprint arXiv:1412.6980  (2014)

\bibitem{sgdepth}
Klingner, M., Term{\"o}hlen, J.A., Mikolajczyk, J., Fingscheidt, T.:
  Self-supervised monocular depth estimation: Solving the dynamic object
  problem by semantic guidance. In: European Conference on Computer Vision. pp.
  582--600. Springer (2020)

\bibitem{kuznietsov2017semi}
Kuznietsov, Y., Stuckler, J., Leibe, B.: Semi-supervised deep learning for
  monocular depth map prediction. In: Proceedings of the IEEE conference on
  computer vision and pattern recognition. pp. 6647--6655 (2017)

\bibitem{instadm}
Lee, S., Im, S., Lin, S., Kweon, I.S.: Learning monocular depth in dynamic
  scenes via instance-aware projection consistency. In: 35th AAAI Conference on
  Artificial Intelligence/33rd Conference on Innovative Applications of
  Artificial Intelligence/11th Symposium on Educational Advances in Artificial
  Intelligence. pp. 1863--1872. ASSOC ADVANCEMENT ARTIFICIAL INTELLIGENCE
  (2021)

\bibitem{lee2021attentive}
Lee, S., Rameau, F., Pan, F., Kweon, I.S.: Attentive and contrastive learning
  for joint depth and motion field estimation. In: Proceedings of the IEEE/CVF
  International Conference on Computer Vision. pp. 4862--4871 (2021)

\bibitem{li2020unsupervised}
Li, H., Gordon, A., Zhao, H., Casser, V., Angelova, A.: Unsupervised monocular
  depth learning in dynamic scenes. In: CoRL (2020)

\bibitem{undeepvo}
Li, R., Wang, S., Long, Z., Gu, D.: Undeepvo: Monocular visual odometry through
  unsupervised deep learning. In: 2018 IEEE international conference on
  robotics and automation (ICRA). pp. 7286--7291. IEEE (2018)

\bibitem{frozen}
Li, Z., Dekel, T., Cole, F., Tucker, R., Snavely, N., Liu, C., Freeman, W.T.:
  Learning the depths of moving people by watching frozen people. In:
  Proceedings of the IEEE/CVF Conference on Computer Vision and Pattern
  Recognition. pp. 4521--4530 (2019)

\bibitem{luo2019every}
Luo, C., Yang, Z., Wang, P., Wang, Y., Xu, W., Nevatia, R., Yuille, A.: Every
  pixel counts++: Joint learning of geometry and motion with {3D} holistic
  understanding. PAMI  (2019)

\bibitem{mahjourian2018unsupervised}
Mahjourian, R., Wicke, M., Angelova, A.: Unsupervised learning of depth and
  ego-motion from monocular video using {3D} geometric constraints. In: CVPR
  (2018)

\bibitem{signet}
Meng, Y., Lu, Y., Raj, A., Sunarjo, S., Guo, R., Javidi, T., Bansal, G.,
  Bharadia, D.: Signet: Semantic instance aided unsupervised 3d geometry
  perception. In: {IEEE} Conference on Computer Vision and Pattern Recognition,
  {CVPR} 2019, Long Beach, CA, USA, June 16-20, 2019. pp. 9810--9820. Computer
  Vision Foundation / {IEEE} (2019). \doi{10.1109/CVPR.2019.01004},
  \url{http://openaccess.thecvf.com/content\_CVPR\_2019/html/Meng\_SIGNet\_Semantic\_Instance\_Aided\_Unsupervised\_3D\_Geometry\_Perception\_CVPR\_2019\_paper.html}

\bibitem{kitti}
Menze, M., Geiger, A.: Object scene flow for autonomous vehicles. In:
  Conference on Computer Vision and Pattern Recognition (CVPR) (2015)

\bibitem{efficientps}
Mohan, R., Valada, A.: Efficientps: Efficient panoptic segmentation.
  International Journal of Computer Vision  \textbf{129}(5),  1551--1579 (2021)

\bibitem{patil2020dont}
Patil, V., Van~Gansbeke, W., Dai, D., Van~Gool, L.: Don’t forget the past:
  Recurrent depth estimation from monocular video. IEEE Robotics and Automation
  Letters  \textbf{5}(4),  6813--6820 (2020)

\bibitem{pillai2018superdepth}
Pillai, S., Ambrus, R., Gaidon, A.: Superdepth: Self-supervised, super-resolved
  monocular depth estimation. In: ICRA (2019)

\bibitem{pilzer2018unsupervised}
Pilzer, A., Xu, D., Puscas, M.M., Ricci, E., Sebe, N.: Unsupervised adversarial
  depth estimation using cycled generative networks. In: 3DV (2018)

\bibitem{poggi20183net}
Poggi, M., Tosi, F., Mattoccia, S.: Learning monocular depth estimation with
  unsupervised trinocular assumptions. In: 3DV (2018)

\bibitem{ranjan2018adversarial}
Ranjan, A., Jampani, V., Kim, K., Sun, D., Wulff, J., Black, M.J.: Competitive
  collaboration: Joint unsupervised learning of depth, camera motion, optical
  flow and motion segmentation. In: CVPR (2019)

\bibitem{unet}
Ronneberger, O., Fischer, P., Brox, T.: U-net: Convolutional networks for
  biomedical image segmentation. In: International Conference on Medical image
  computing and computer-assisted intervention. pp. 234--241. Springer (2015)

\bibitem{featdepth}
Shu, C., Yu, K., Duan, Z., Yang, K.: Feature-metric loss for self-supervised
  learning of depth and egomotion. In: European Conference on Computer Vision.
  pp. 572--588. Springer (2020)

\bibitem{monoresmatch}
Tosi, F., Aleotti, F., Poggi, M., Mattoccia, S.: Learning monocular depth
  estimation infusing traditional stereo knowledge. In: {IEEE} Conference on
  Computer Vision and Pattern Recognition, {CVPR} 2019, Long Beach, CA, USA,
  June 16-20, 2019. pp. 9799--9809. Computer Vision Foundation / {IEEE} (2019).
  \doi{10.1109/CVPR.2019.01003},
  \url{http://openaccess.thecvf.com/content\_CVPR\_2019/html/Tosi\_Learning\_Monocular\_Depth\_Estimation\_Infusing\_Traditional\_Stereo\_Knowledge\_CVPR\_2019\_paper.html}

\bibitem{kitti2017}
Uhrig, J., Schneider, N., Schneider, L., Franke, U., Brox, T., Geiger, A.:
  Sparsity invariant cnns. In: International Conference on 3D Vision (3DV)
  (2017)

\bibitem{ddvo}
Wang, C., Buenaposada, J.M., Zhu, R., Lucey, S.: Learning depth from monocular
  videos using direct methods. In: Proceedings of the IEEE conference on
  computer vision and pattern recognition. pp. 2022--2030 (2018)

\bibitem{wang2017learning}
Wang, C., Buenaposada, J.M., Zhu, R., Lucey, S.: Learning depth from monocular
  videos using direct methods. In: CVPR (2018)

\bibitem{wang2020self}
Wang, J., Zhang, G., Wu, Z., Li, X., Liu, L.: Self-supervised joint learning
  framework of depth estimation via implicit cues. arXiv:2006.09876  (2020)

\bibitem{wang2019recurrent}
Wang, R., Pizer, S.M., Frahm, J.M.: Recurrent neural network for (un-)
  supervised learning of monocular video visual odometry and depth. In:
  Proceedings of the IEEE/CVF Conference on Computer Vision and Pattern
  Recognition. pp. 5555--5564 (2019)

\bibitem{ssim}
Wang, Z., Bovik, A.C., Sheikh, H.R., Simoncelli, E.P.: Image quality
  assessment: from error visibility to structural similarity. IEEE transactions
  on image processing  \textbf{13}(4),  600--612 (2004)

\bibitem{depthhint}
Watson, J., Firman, M., Brostow, G.J., Turmukhambetov, D.: Self-supervised
  monocular depth hints. In: 2019 {IEEE/CVF} International Conference on
  Computer Vision, {ICCV} 2019, Seoul, Korea (South), October 27 - November 2,
  2019. pp. 2162--2171. {IEEE} (2019). \doi{10.1109/ICCV.2019.00225},
  \url{https://doi.org/10.1109/ICCV.2019.00225}

\bibitem{manydepth}
Watson, J., Mac~Aodha, O., Prisacariu, V., Brostow, G., Firman, M.: The
  temporal opportunist: Self-supervised multi-frame monocular depth. In:
  Proceedings of the IEEE/CVF Conference on Computer Vision and Pattern
  Recognition. pp. 1164--1174 (2021)

\bibitem{monorec}
Wimbauer, F., Yang, N., Von~Stumberg, L., Zeller, N., Cremers, D.: Monorec:
  Semi-supervised dense reconstruction in dynamic environments from a single
  moving camera. In: Proceedings of the IEEE/CVF Conference on Computer Vision
  and Pattern Recognition. pp. 6112--6122 (2021)

\bibitem{monodispnet}
Wong, A., Soatto, S.: Bilateral cyclic constraint and adaptive regularization
  for unsupervised monocular depth prediction. In: {IEEE} Conference on
  Computer Vision and Pattern Recognition, {CVPR} 2019, Long Beach, CA, USA,
  June 16-20, 2019. pp. 5644--5653. Computer Vision Foundation / {IEEE} (2019).
  \doi{10.1109/CVPR.2019.00579},
  \url{http://openaccess.thecvf.com/content\_CVPR\_2019/html/Wong\_Bilateral\_Cyclic\_Constraint\_and\_Adaptive\_Regularization\_for\_Unsupervised\_Monocular\_Depth\_CVPR\_2019\_paper.html}

\bibitem{lego}
Yang, Z., Wang, P., Wang, Y., Xu, W., Nevatia, R.: {LEGO:} learning edge with
  geometry all at once by watching videos. In: 2018 {IEEE} Conference on
  Computer Vision and Pattern Recognition, {CVPR} 2018, Salt Lake City, UT,
  USA, June 18-22, 2018. pp. 225--234. {IEEE} Computer Society (2018).
  \doi{10.1109/CVPR.2018.00031},
  \url{http://openaccess.thecvf.com/content\_cvpr\_2018/html/Yang\_LEGO\_Learning\_Edge\_CVPR\_2018\_paper.html}

\bibitem{geonet2018}
Yin, Z., Shi, J.: {GeoNet}: Unsupervised learning of dense depth, optical flow
  and camera pose. In: CVPR (2018)

\bibitem{zhanst2018}
Zhan, H., Garg, R., Weerasekera, C.S., Li, K., Agarwal, H., Reid, I.:
  Unsupervised learning of monocular depth estimation and visual odometry with
  deep feature reconstruction. In: CVPR (2018)

\bibitem{dfr}
Zhan, H., Garg, R., Weerasekera, C.S., Li, K., Agarwal, H., Reid, I.D.:
  Unsupervised learning of monocular depth estimation and visual odometry with
  deep feature reconstruction. In: 2018 {IEEE} Conference on Computer Vision
  and Pattern Recognition, {CVPR} 2018, Salt Lake City, UT, USA, June 18-22,
  2018. pp. 340--349. {IEEE} Computer Society (2018).
  \doi{10.1109/CVPR.2018.00043},
  \url{http://openaccess.thecvf.com/content\_cvpr\_2018/html/Zhan\_Unsupervised\_Learning\_of\_CVPR\_2018\_paper.html}

\bibitem{zhang2019exploiting}
Zhang, H., Shen, C., Li, Y., Cao, Y., Liu, Y., Yan, Y.: Exploiting temporal
  consistency for real-time video depth estimation. In: Proceedings of the
  IEEE/CVF International Conference on Computer Vision. pp. 1725--1734 (2019)

\bibitem{peloss}
Zhao, H., Gallo, O., Frosio, I., Kautz, J.: Loss functions for image
  restoration with neural networks. IEEE Transactions on computational imaging
  \textbf{3}(1),  47--57 (2016)

\bibitem{dualnet}
Zhou, J., Wang, Y., Qin, K., Zeng, W.: Unsupervised high-resolution depth
  learning from videos with dual networks. In: 2019 {IEEE/CVF} International
  Conference on Computer Vision, {ICCV} 2019, Seoul, Korea (South), October 27
  - November 2, 2019. pp. 6871--6880. {IEEE} (2019).
  \doi{10.1109/ICCV.2019.00697}, \url{https://doi.org/10.1109/ICCV.2019.00697}

\bibitem{zhou2017unsupervised}
Zhou, T., Brown, M., Snavely, N., Lowe, D.: Unsupervised learning of depth and
  ego-motion from video. In: CVPR (2017)

\bibitem{dfnet}
Zou, Y., Luo, Z., Huang, J.B.: Df-net: Unsupervised joint learning of depth and
  flow using cross-task consistency. In: Proceedings of the European conference
  on computer vision (ECCV). pp. 36--53 (2018)

\end{thebibliography}
\end{document}